\documentclass[10pt,twocolumn,letterpaper]{article}

\usepackage{cvpr}              

\usepackage{graphicx}
\usepackage{amsmath}
\usepackage{amssymb}
\usepackage{booktabs}
\usepackage{multirow}
\usepackage{multicol}

\usepackage[pagebackref,breaklinks,colorlinks]{hyperref}

\usepackage[capitalize]{cleveref}
\crefname{section}{Sec.}{Secs.}
\Crefname{section}{Section}{Sections}
\Crefname{table}{Table}{Tables}
\crefname{table}{Tab.}{Tabs.}

\usepackage{color}

\definecolor{cb_orange}{rgb}{1.0,0.51,0.0}
\definecolor{cb_blue}{rgb}{0.22,0.49,0.72}
\definecolor{cb_green}{rgb}{0.3,0.67,0.29}
\definecolor{cb_red}{rgb}{0.89,0.1,0.11}
\definecolor{cb_pink}{rgb}{1, 0, 0.4}

\def\cvprPaperID{*****} 
\def\confName{CVPR}
\def\confYear{2023}

\begin{document}

\title{I2V: Towards Texture-Aware Self-Supervised Blind Denoising using Self-Residual Learning for Real-World Images}

\author{Kanggeun~Lee \hspace{1cm} Kyungryun~Lee  \hspace{1cm}  Won-Ki~Jeong\\
Department of Computer Science and Engineering, Korea University \\
{\tt\small leekanggeun@gmail.com, \{krlee0000, wkjeong\}@korea.ac.kr}
}
\maketitle

\begin{abstract}
Although the advances of self-supervised blind denoising are significantly superior to conventional approaches without clean supervision in synthetic noise scenarios, it shows poor quality in real-world images due to spatially correlated noise corruption. 
Recently, pixel-shuffle downsampling (PD) has been proposed to eliminate the spatial correlation of noise. A study combining a blind spot network (BSN) and asymmetric PD (AP) successfully demonstrated that self-supervised blind denoising is applicable to real-world noisy images. 
However, PD-based inference may degrade texture details in the testing phase because high-frequency details (e.g., edges) are destroyed in the downsampled images.
To avoid such an issue, we propose self-residual learning without the PD process to maintain texture information. We also propose an order-variant PD constraint, noise prior loss, and an efficient inference scheme (progressive random-replacing refinement ($\text{PR}^3$)) to boost overall performance. The results of extensive experiments show that the proposed method outperforms state-of-the-art self-supervised blind denoising approaches, including several supervised learning methods, in terms of PSNR, SSIM, LPIPS, and DISTS in real-world sRGB images.
\end{abstract}


\begin{figure}[t]
\includegraphics[width=8cm,keepaspectratio]{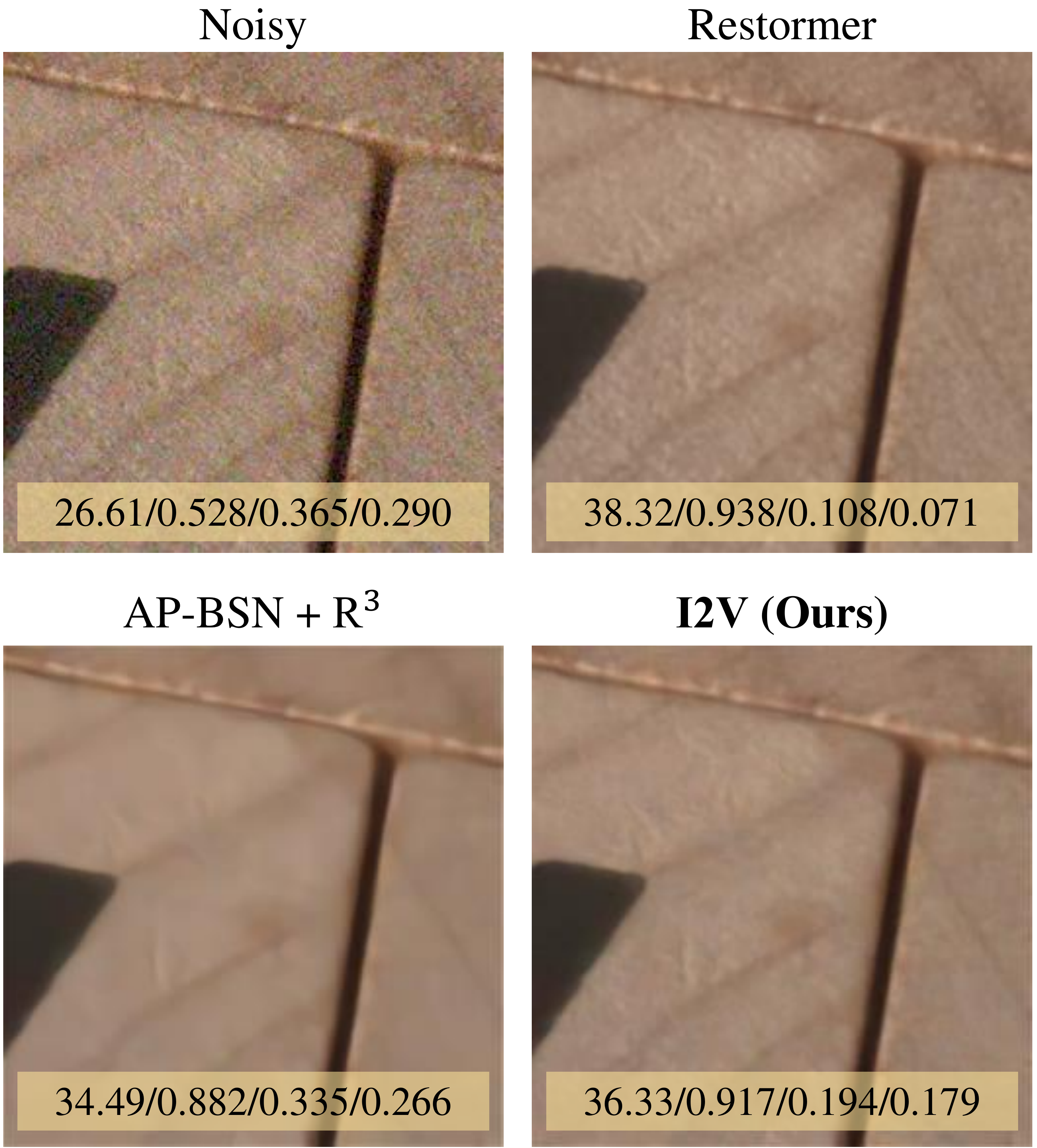}
\caption{Example of real-world image denoising in SIDD validation dataset. Restormer~\cite{restormer} is a supervised learning-based denoising method. AP-BSN + $\text{R}^3$~\cite{apbsn} and our method (\textbf{I2V}) are self-supervised blind denoising approaches. From left to right: PSNR (peak signal-to-noise ratio) $\uparrow$ / SSIM (structural similarity) $\uparrow$~\cite{ssim} / LPIPS (learned perceptual image patch similarity) $\downarrow$~\cite{lpips} / DISTS (deep image structure and texture similarity) $\downarrow$~\cite{dists}.}
   \label{fig:fig1}
\end{figure}

\section{Introduction}
\label{sec:intro}
Image denoising is a low-level computer vision 
problem for 
restoring a clean image from its noisy observation. 
%
%
Unlike conventional approaches relying on image priors (e.g., sparse representation~\cite{elad2006image}, total variation~\cite{vese2003modeling}, and non-local self-similarity~\cite{bm3d,wnnm}), the advances of convolutional neural network (CNN) architectures~\cite{zhang2017beyond,redcnn,nafnet,restormer} have afforded superior denoising performance using clean-noisy training pairs. 
Despite the superior denoising performance, these data-driven approaches suffer from a lack of sufficient clean-noisy pairs for training deep neural networks, thus hindering their widespread application in real-world scenarios, where matching clean images are unavailable. 
%
Recently, self-supervised learning-based denoisers~\cite{noise2void,noise2self,self2self,neighbor2neighbor} have shown promising denoising performance with only noisy observations without clean images nor noise statistics. 
The $\mathcal{J}$-$invariant$ property~\cite{noise2self} enabled self-supervised learning to be considered as supervised learning under the assumptions of zero-mean noise and pixel-wise signal-independent noise.
However, most existing methods have been tested only in a controlled setup and are known to perform poorly on real-world sRGB noisy images, such as SIDD~\cite{SIDD}, DND~\cite{dnd}, and NIND~\cite{nind}, due to spatial correlation of camera noise. 
%
%

Recently, Zhou~\textit{et al.}~\cite{zhouetal} proposed pixel-shuffle downsampling (PD) to break spatially-correlated real noise into pixel-wise independent noise, making it feasible to train the real-world image denoiser using synthetic data with additive white Gaussian noise (AWGN). 
Later, Lee~\textit{et al.}~\cite{apbsn} proposed AP-BSN, which extends PD to asymmetric PD (AP) using a large PD stride factor ($s=5$) for training to ensure pixel-wise independent constraint and a small PD stride factor ($s=2$) for inference to maximize reconstruction quality so that fully self-supervised training of a denoiser using a blind spot network (BSN)~\cite{DBSN} is feasible on real-world images. 
Although employing AP was an appropriate choice for integrating PD and BSN, we observed some issues with its image quality. 
As shown in Figure~\ref{fig:fig1}, the AP-BSN result shows excessive blurring and loss of texture details. 
We identified that this is mainly due to downsampling in PD.  
Because the stride factor of 5 in AP-BSN converts the input noisy image into extremely small images (1/25 of the input size), the deep learning model only sees highly corrupted low-resolution images during training, and there is no chance that the model learns high-frequency details in the original resolution of the image. 
%
AP-BSN compensates for this by using the minimum stride factor ($s=2$) during the test, but information loss persists even in quarter-sized images (in fact, this is a common problem in PD~\cite{zhouetal} as well). 
Moreover, because matching the distribution between the training and test data is one of the critical issues associated with real-world machine learning applications~\cite{gcbd,c2n,DBSN}, asymmetric image scales in AP-BSN introduce further performance issues.

In this paper, we propose~\textit{Invariant2Variant (I2V)}, a novel fully self-supervised blind denoising framework that overcomes the limitations of the PD process and mismatching data distributions in AP-BSN, specifically aiming to improve texture details in the denoised result. 
I2V leverages self-residual learning~\cite{iscl} over self-supervised learning to unify the distributions of the training and test data. 
%
Herein we also propose a novel order-variant PD, which is inspired by Nei2Nei~\cite{neighbor2neighbor}, for training data augmentation and content similarity loss. 
%
%
To prevent an overfitting issue in learning with pseudo-noise labels, we also propose a noise prior loss as a regularizer.
%
Finally, we propose a new inference scheme, progressive random-replacing refinement ($\text{PR}^3$), that does not require PD downsampling for inference. 
We demonstrate that I2V outperforms state-of-the-art self-supervised blind denoisers on real-world images in terms of various image quality metrics, including PSNR, SSIM, LPIPS~\cite{lpips}, and DISTS~\cite{dists}. 
%
Our contributions can be summarized as follows: 

\begin{itemize}
\item We propose a novel real-world image denoiser based on both self-supervised and self-residual learning. By using the proposed method, we can reduce the data distribution mismatch issue and improve the texture details in the denoised result. 
%
\item We propose a novel order-variant PD process for training data augmentation. 
%
We also propose a novel inference scheme, \textbf{$\text{PR}^{3}$}, that reduces running time while improving image quality.
\item We demonstrate that the proposed denoiser can effectively preserve texture details even better than supervised learning methods for some cases, assessed by the perception-based image quality metrics (LPIPS and DISTS).


%
\end{itemize}

\begin{figure*}[t]
\centering
\includegraphics[width=16.5cm,keepaspectratio]{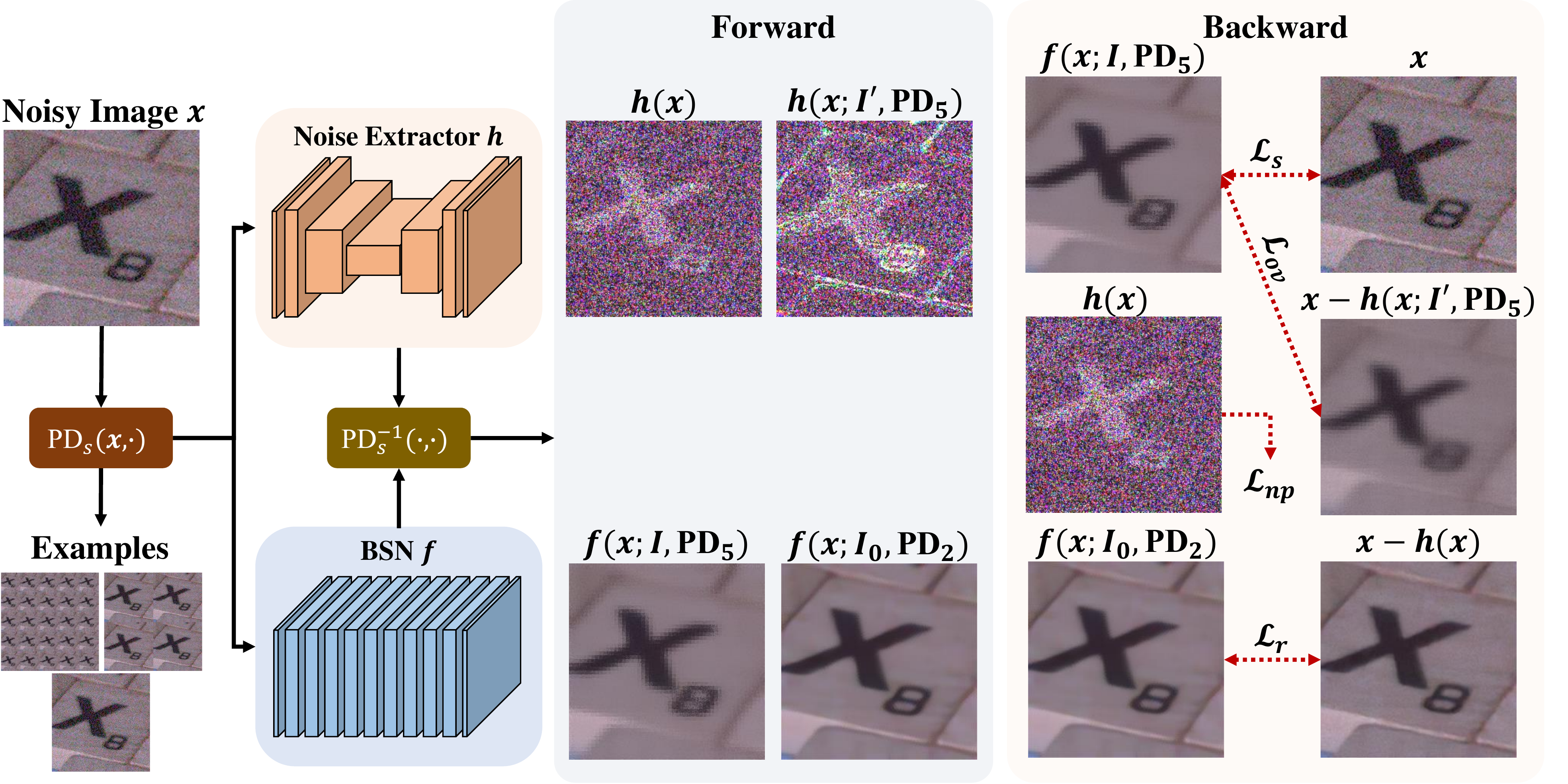}
\caption{Overview of the training scheme in our proposed \textbf{I2V} framework. Examples show three cases for $\text{PD}_5(x,\cdot)$, $\text{PD}_2(x,\cdot)$, and $\text{PD}_1(x,\cdot)=x$. $f(x;I,\text{PD}_s)$ is defined as $\text{PD}^{-1}_s(f(\text{PD}_s(x,I)),I^T)$. $I_0$ is the identity transformation matrix, and $I$ and $I'$ are random transformation matrices. Two networks $f$ and $h$ are trained using four loss functions: $\mathcal{L}_\text{s}$, $\mathcal{L}_{\text{r}}$, $\mathcal{L}_{\text{ov}}$, and $\mathcal{L}_{\text{np}}.$}
   \label{fig:fig2}
\end{figure*}

\section{Related Work}
\textbf{Supervised denoising.} Recently, deep learning-based denoisers~\cite{zhang2017beyond, ffdnet, redcnn} showed superior performance over traditional algorithms~\cite{chambolle2004algorithm,bm3d,wnnm,vese2003modeling,elad2006image} in simulated data with specific noise statistics, such as AWGN.
%
%
Nonetheless, these methods trained by synthetic clean-noisy pairs cause performance degradation~\cite{cbdnet} in real-world noisy images that belong to a different distribution. With the advances of deep learning model architectures~\cite{nafnet,restormer} and training strategy~\cite{danet}, the clean-noisy pairs in current real-world scenarios can lead to optimal performance in the target distribution. However, collecting clean-noisy pairs is a critical limitation, which is sometimes infeasible in practice.
Therefore, most current studies are evolving to self-supervised blind denoising without noise statistics and clean images.

\textbf{Unpaired image denoising.} GCBD~\cite{gcbd} showed the possibility of training from unpaired clean-noisy images through a GAN\cite{gan}, and further studies~\cite{DBSN,hong2020end,c2n,iscl} have achieved performances close to that of supervised learning. 
ISCL~\cite{iscl} leverages self-residual learning with cycle-GAN~\cite{cyclegan} to overcome a self-adversarial attack~\cite{selfadvatt} that causes a performance decrease. As unpaired image denoising takes advantage of clean images, it performs relatively better than self-supervised denoising, which exploits noisy images only; nevertheless, the clean images collected should have similar statistics to the target distribution for denoising, which is also labor-intensive in certain domains. 

\textbf{Blind denoising without clean supervision.} Lehtinen \textit{et al.}~\cite{noise2noise} first introduced the N2N paradigm, only requiring noisy image pairs for training a deep neural network. 
The latest research has proposed various self-supervised denoising methods in which prior knowledge of noise is necessary ~\cite{noisier2noise, r2r, Kim_2022_CVPR} or not~\cite{noise2void, noise2self, self2self, cvfsid}.  
Moreover,~\cite{DBSN, noise2kernel} derived dilated convolution layers with a single donut kernel-based layer that always satisfy the $\mathcal{J}$-$invariant$ property. 
Recently, Nei2Nei~\cite{neighbor2neighbor} proposed creating noisy image pairs by randomly sub-sampling neighbor pixels to utilize the assumption of N2N.
Although showing remarkable denoising results in raw-RGB images,
it failed to deal with sRGB noisy images, as shown in our manuscript.
To apply such self-supervised denoising approaches, noise statistics must satisfy pixel-wise independent and zero-mean noise assumptions; however, real-world noise consists of various structured patterns that are spatially correlated, thus violating these assumptions. 
To handle spatially correlated noise in the sRGB space, the PD process was proposed to break spatial correlation ~\cite{zhouetal,apbsn} such that self-supervised denoising can be applied to spatially invariant noise in real-world images. 
%




\section{Method}
In this section, we introduce the details of I2V including the proposed loss functions and inference scheme ($\text{PR}^3$). 
In Figure~\ref{fig:fig2}, we employ an arbitrary function $h$ as a noise extractor. 
In the forward pass, four outputs will be generated by $f$ and $h$ for various stride factors (s=1,2,5). 
We optimize $f$ and $h$ using the loss functions introduced in the following sections.
%
%
%
%
\begin{figure}[t]
\centering
\includegraphics[width=6.5cm,keepaspectratio]{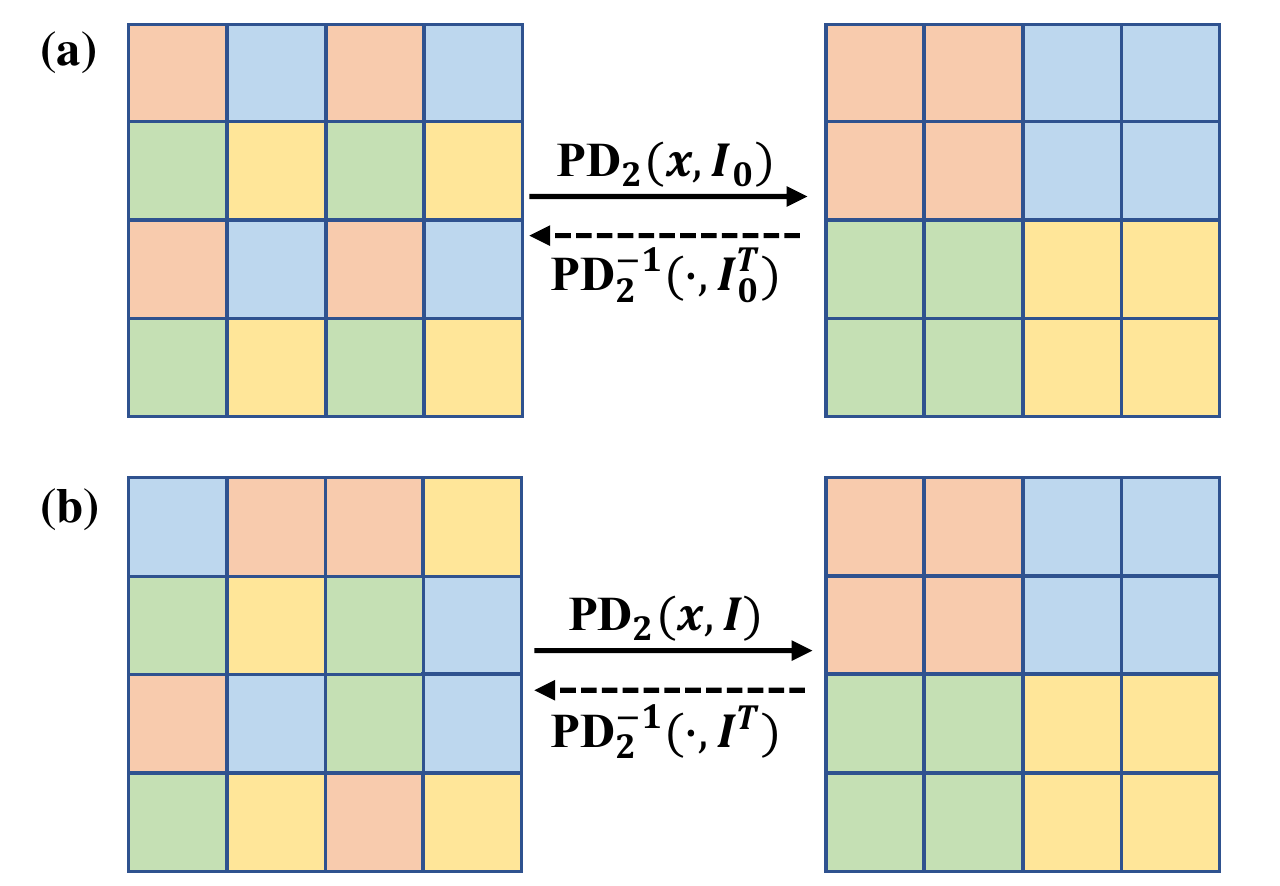}
\caption{Examples of order-invariant and order-variant PD for stride factor 2. (a) Original PD~\cite{zhouetal} (order-invariant) with an identity transformation matrix $I_0$. (b) The proposed order-variant PD with a randomly generated transformation matrix $I\in \mathcal{I}$.}
   \label{fig:fig3}
\end{figure}

\begin{figure*}[t]
\centering
\includegraphics[width=16cm,keepaspectratio]{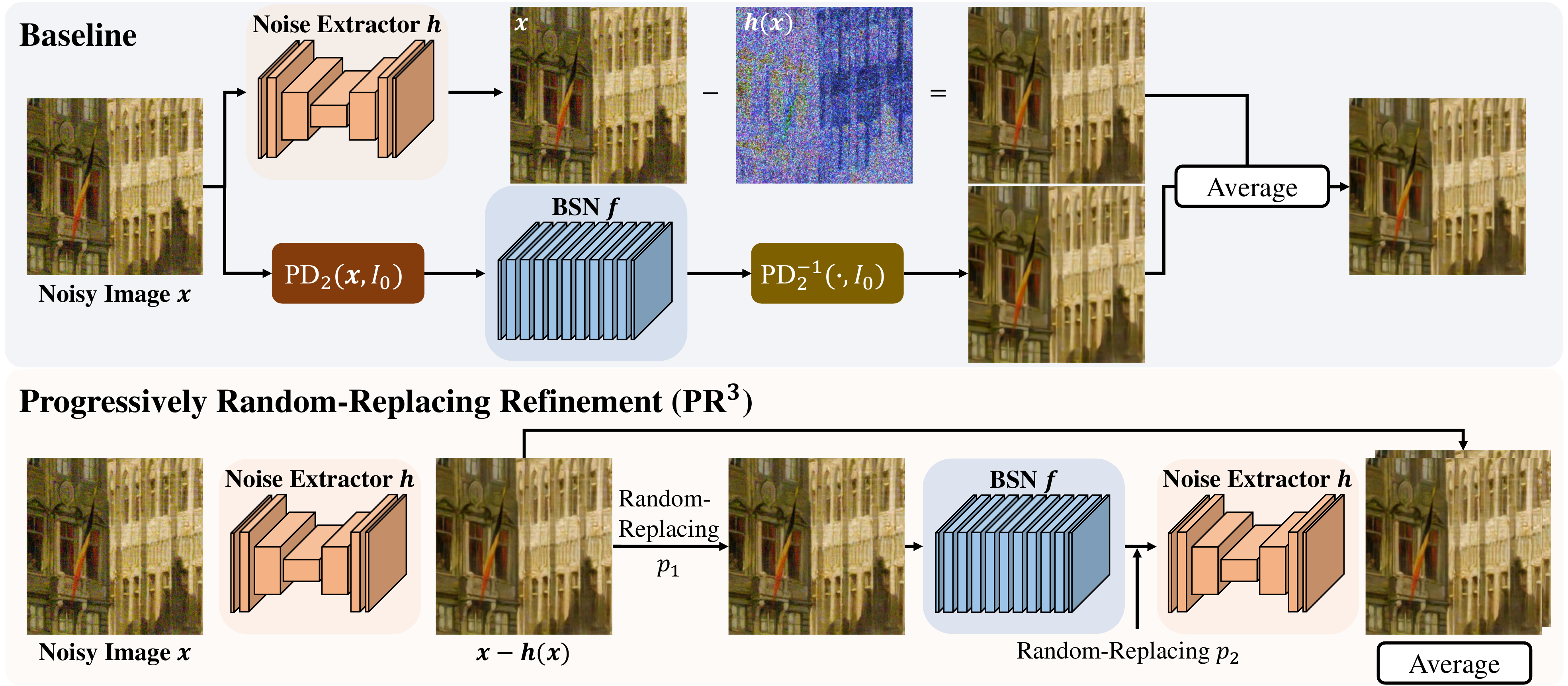}
\caption{Overview of the baseline and $\text{PR}^3$ inference strategies. Unlike BSN $f$, the noise extractor $h$ does not require $\text{PD}_s$ to satisfy the input spatially uncorrelated noise condition. 
In $\text{PR}^3$, random-replacing makes spatially uncorrelated pixel-wise independent noise, so $\text{PD}_s$ is not required even for $f$.} 
   \label{fig:fig4}
\end{figure*}

\subsection{AP-BSN Revisit}
\label{sec:apbsn_revisit}

AP-BSN is a variant of BSN using different PD stride factors in training and test phases. 
%
%
%
Let us define $f(x;I,\text{PD}_s) := \text{PD}^{-1}_{s}(f(\text{PD}_{s}(x,I)),I^T)$ with $I\in{\mathcal{I}}$, which is a transformation matrix to restore the original image from its pixel-shuffle down-sampled image, as shown in Figure~\ref{fig:fig3}. 
Then, we reformulate the self-supervised loss of AP-BSN as follows:
\begin{align}
    \mathcal{L}_{\text{s}}(f,\mathcal{X}) =
    \mathbb{E}_{x}||(f(x;I_0,\text{PD}_5)-x||_1
    \label{eq:eq1}
\end{align}
where $x\in \mathcal{X}$ is a noisy image, $f$ is the BSN~\cite{DBSN}, $\text{PD}_s$ is a PD function with stride factor $s$, and $I_0$ is an identity transformation matrix for \textit{order-invariant} PD, as shown in Figure~\ref{fig:fig3}. 
%
Note that because order-invariant PD generates down-sampled images via a pre-defined pixel-sampling order, a limited number of sub-images are generated. 
For example, Figure~\ref{fig:fig3} (a) shows that order-invariant PD generates four $2 \times 2$ sub-images from a $4 \times 4$ input image. 
To increase the number of sub-images, we propose \textit{order-variant} PD using a randomly chosen transformation matrix $I$ that shuffles the sampling order (see Figure~\ref{fig:fig3} (b)). 
Then, Eq.~\ref{eq:eq1} 
can be reformulated as follows:
%
\begin{align}
    \mathcal{L}_{\text{s}}(f,\mathcal{X},\mathcal{I}) =
    \mathbb{E}_{x,I\in\mathcal{I}}||(f(x;I,\text{PD}_5)-x||_1
    \label{eq:eq2}
\end{align}

To further boost the denoising performance and reduce the visual artifacts~\cite{zhouetal} of AP-BSN, post-processing called random-replacing refinement ($\text{R}^3$) is proposed. 
%
%
%
$\text{R}^3$ represents an average of restored images from multiple synthetic noisy images, which are generated by randomly replacing pixel-wise noise (i.e., selected from $x$) into the initial prediction $f(x;I_0,PD_2)$. 
We observed that $\text{R}^3$ tends to over-smooth the restored image through repeated noise removal, which causes the loss of texture details. 
We address this issue in Section~\ref{sec:pr3}.

%
%
%
\subsection{Self-Residual Learning}
In this section, we introduce a novel loss function to address the issues of training-inference data distribution mismatch and blurring artifacts due to excessive downsampling of $\text{PD}_5$ during training. 
%
The main idea is that if $\text{PD}_2$ down-sampled images are used in the inference phase, then they 
should be used during the training phase as well. 
For this, we introduce a noise extractor network $h$ trained in a self-supervised manner using a pseudo-noise map $x-f(x,I_0,\text{PD}_2)$ for residual (noise) learning. 
The corresponding self-residual loss is defined as follows:
\begin{align}
    \mathcal{L}_{\text{r}}(f,h,\mathcal{X}) = \mathbb{E}_{x}||x-f(x;I_0,\text{PD}_2)-h(x)||_1
    \label{eq:eq3}
\end{align}
where the order-invariant PD with $I_0$ is used to minimize the aliasing artifacts for training $h$. 
Our interpretation of this loss is as follows.
The pseudo-noise map $x-f(x;I_0,\text{PD}_2)$ may include two kinds of noises: spatially-correlated real noise and aliasing artifacts from downsampling. 
%
%
We observed that $x-h(x)$ shows higher texture restoration quality compared to $f(x;I_0,\text{PD}_2)$ (see supplementary material).
Another benefit is that, unlike BSN that is trained using only $\text{PD}_5$, the training data for the noise extractor consist of high-resolution noisy images only. 
%
%
%
%
Note also that the network structure of $h$ does not require the $\mathcal{J}$-$invariant$ property; thus, any state-of-the-art image restoration network architectures (such as~\cite{nafnet, restormer}) can be employed for $h$. 
%

\subsection{Order-Variant PD Constraint}
In this section, we propose another loss function designed to promote content (low-frequency features) similarity between two predicted images, as shown by $f(x;\cdot,\text{PD}_5)$ and $x-h(x;\cdot,\text{PD}_5)$ in Figure~\ref{fig:fig2}. 
Order-variant $\text{PD}_5$ with a random transformation increases aliasing artifacts, which can be considered as spatially uncorrelated, pixel-wise independent noise. 
Therefore, by applying $f$ and $h$ for a given noisy image $x$, its aliasing artifacts will be effectively removed and only low-frequency content information will remain.
%
%
%
%
%
Based on this observation, we propose an order-variant PD constraint loss as follows:
\begin{align}
    \nonumber
    \mathcal{L}_{\text{ov}}(&f,h,\mathcal{X},\mathcal{I}) = \\ &\mathbb{E}_{x;I,I'\in\mathcal{I}}||x-f(x;I,PD_5)-h(x;I',PD_5)||_1
    \label{eq:eq5}
\end{align}
where $I$ and $I'$ are random transformation matrices.
This loss term enforces the content information of predictions from $f$ and $h$ to be close each other. 
Therefore, this loss contributes to the overall shape and content restoration and improves PSNR and SSIM.


%
%
%
\subsection{Noise Prior Constraint}

Although the proposed noise extractor $h$ improves denoising quality by matching data distribution, we still observed some loss of texture details and color shifts, especially in the texture-rich images.
%
%
We observed that this is because the noise extractor $h$ overfits the aliasing artifacts of $\text{PD}_2$ as well as real noise in $\mathcal{L}_{\text{r}}$  
(see the supplemental Figure S1 showing color shifts as well as texture deformations in the AP-BSN prediction).  
As shown in the supplemental Figure S2, such aliasing artifacts contribute to texture details and their magnitude is larger than that of real noise. 
To further improve the texture details, we propose another loss function, \textit{noise prior loss}, that limits the distribution of $h(x)$ (i.e., penalizing high magnitude noises) using the following $\text{L}_1$-regularization term as follows:
\begin{align}
    \mathcal{L}_{\text{np}}(h,\mathcal{X}) = \mathbb{E}_{x}||\mathbb{E}_{j\in J}[h(x)_{j}]||_1,
    \label{eq:eq6}
\end{align}
where $J$ is a set of indices to indicate mini-batch and color axes. 
This regularization term takes 
the pixel-wise absolute value of the mean along mini-batch and color axes, making the outliers of the pseudo-noise map in Eq.~\ref{eq:eq3} effectively suppressed.  
%

\vspace{0.05in}
\noindent \textbf{Full objective.} Finally, we propose a total loss for BSN $f$ and the noise extractor $h$ as follows:
\begin{align}
\nonumber
    \mathcal{L}_{\text{total}} &= \lambda_{\text{s}} \mathcal{L}_{\text{s}}(f,\mathcal{X},\mathcal{I}) + \lambda_{\text{r}} \mathcal{L}_{\text{r}}(f,h,\mathcal{X}) \\
    &+ \lambda_{\text{ov}} \mathcal{L}_{\text{ov}}(f,h,\mathcal{X},\mathcal{I}) + \lambda_{\text{np}} \mathcal{L}_{\text{np}}(h,\mathcal{X})
    \label{eq:eq7}
\end{align}
where the hyperparameters $\lambda_{\text{s}}, \lambda_{\text{r}}, \lambda_{\text{ov}}$, and $\lambda_{\text{np}}$ imply the contribution weight of each loss.

\subsection{Progressive Random-Replacing Refinement}
\label{sec:pr3}
Even after minimizing the total loss, visual artifacts may still remain in the  baseline inference result due to the structural limitation of BSN $f$ and $\text{PD}_2$. 
The random-replacing refinement ($\text{R}^3$) strategy proposed in AP-BSN is a powerful tool to mitigate visual artifacts. 
However, averaging multiple predictions for various noisy samples increases content similarity and decreases texture details. 
Moreover, the baseline of $\text{R}^3$ relies on the initial prediction that the texture details are degraded by $\text{PD}_2$ downsampling. 
To address these drawbacks of $\text{R}^3$, we propose a $\text{PR}^3$, as shown in Figure~\ref{fig:fig4}.
%
We define the random-replacing function $g$ as follows:
\begin{align}
    g(\mathcal{M},x,y') = \mathcal{M}\odot {x} + (1-\mathcal{M})\odot {y'}
    \label{eq:eq8}
\end{align}
where $\odot$ is the Hadamard product, $\mathcal{M}$ is a binary mask, and $y'$ is a denoised image from any function. The binary mask $\mathcal{M}\in{\{0,1\}^{C\times H\times W}}$ denotes the matrix that is independently sampled from a Bernoulli distribution with probability $p\in(0,1)$.
To maintain the texture details of the original image, we set an initial prediction $\hat{y}:=x-h(x)$ of $h$ as the primary result. 
Then, progressively denoised predictions $\tilde{y}_{\text{BSN}}$ and $\tilde{y}_{\text{NE}}$ with the function $g$ are generated as follows:
\begin{align}
    \tilde{y}_{\text{BSN}} &= f(g(\mathcal{M}_1,x,\hat{y})) \\
    \tilde{n}_{\text{NE}} &= h(g(\mathcal{M}_2,x,\tilde{y}_{\text{BSN}})) \\
    \tilde{y}_{\text{NE}} &= (1-\mathcal{M}_2)\odot \tilde{y}_{\text{BSN}} + \mathcal{M}_2\odot (x-\tilde{n}_{\text{NE}})
    \label{eq:eq11}
\end{align}
where binary masks ($\mathcal{M}_1$ and $\mathcal{M}_2$) are generated by $p_1$ and $p_2$, respectively.
Finally, the average of the primary result $\hat{y}$ and the last denoised image $\tilde{y}_{\text{NE}}$ will be the final prediction output.
\begin{table*}
  \centering
  \renewcommand{\arraystretch}{1.2}
  \renewcommand{\tabcolsep}{2pt}
  \begin{tabular}{c|c|c|c|c|c|c|c|c|c}
    \hline
    \multirow{2}{*}{Learning Type} & \multirow{2}{*}{Method} & \multicolumn{4}{c|}{SIDD Validation} & \multicolumn{2}{c|}{SIDD Benchmark} & \multicolumn{2}{c}{DND Benchmark} \\ 
                                   &                          & PSNR $\uparrow$ & SSIM $\uparrow$ & LPIPS $\downarrow$ & DISTS $\downarrow$ & PSNR $\uparrow$ & SSIM $\uparrow$ & PSNR $\uparrow$ & SSIM $\uparrow$ \\ \hline
    
    \multirow{4}{*}{Supervised}  & DnCNN~\cite{zhang2017beyond}  & 35.25  & 0.861 & 0.272 & 0.218 & 35.25 & 0.905                              & 37.61 & 0.934 \\
                                 & DnCNN$^\dagger$~\cite{zhouetal} & 35.45 & 0.885 & 0.288 & 0.232 & 35.44 & 0.924 & 37.79 & 0.940 \\
                                 & DANet~\cite{danet}  & 39.00 & 0.914 & 0.263 & 0.226 & 38.89 & 0.952 & 39.13 & 0.948\\
                                 & NAFNet~\cite{nafnet} & 39.37  & 0.918 & 0.249 & 0.218 & 39.26 & 0.956 & 39.12 & 0.949 \\
                                 & Restormer~\cite{restormer} & 39.02  & 0.914 & 0.221 & 0.191  & 38.89 & 0.953 & 39.22 & 0.949 \\
                                 \hline 
    \multirow{1}{*}{Unpaired image-based} 
                                          & C2N~\cite{c2n}+DIDN~\cite{didn} & 35.39 & 0.891 & 0.237 & 0.199 & 35.35 & 0.930 & 38.14 & 0.941\\ \hline
                                          
    \multirow{6}{*}{Self-supervised}& N2V~\cite{noise2void} & 27.06 & 0.551 & 0.468 & 0.332 &                                 26.99 & 0.652 & 29.23 & 0.765\\ 
                                    & Nei2Nei~\cite{neighbor2neighbor} & 27.94 & 0.604 & 0.441 & 0.317 & 27.90 & 0.679 & 30.87 & 0.792\\ 
                                    & AP-BSN~\cite{apbsn}  & 36.23 & 0.853 & 0.281 & 0.255 & 36.19 & 0.913 & 37.73 & 0.928\\ 
                                    & AP-BSN + $\text{R}^3$~\cite{apbsn} & 36.30 & \underline{\textbf{0.890}} & 0.315 & 0.267 & 36.19 & 0.927 & 37.00 & 0.934 \\ 
                                    & $\text{I2V}^{\textbf{\text{B}}}$ (Ours)               & \underline{\textbf{36.63}} & 0.888  & \underline{0.251}    &    \underline{0.218}    & \underline{\textbf{36.52}} & \underline{\textbf{0.931}} & \underline{\textbf{38.08}}  & \underline{0.938} \\ 
                                    & I2V (Ours) & \underline{36.48} & \underline{0.889} & \underline{\textbf{0.245}} & \underline{\textbf{0.199}} & \underline{36.35} & \underline{0.929} & \underline{37.87}  & \underline{\textbf{0.939}}  \\ \hline
  \end{tabular}
  \caption{Quantitative results on the SIDD validation, SIDD benchmark, and DND benchmark datasets. Supervised denoising and unpaired image denoising approaches leverage paired clean-noisy images while self-supervised learning methods rely on only noisy images in SIDD-Medium dataset. $\dagger$ indicates a trained network by synthetic noise (AWGN, random value impulse noise) with PD refinement. $\text{I2V}^{\textbf{\text{B}}}$ represents I2V with the baseline inference scheme in place of $\text{PR}^3$. The best and second-best are underlined, and the best results are marked in bold among self-supervised learning methods.}
  \label{tab:tab1}
\end{table*}

\section{Experiment}
\subsection{Implementation Details}
\noindent \textbf{Training details.} 
We implemented I2V using Pytorch 1.12.0~\cite{pytorch}. 
We used RAdam optimizer~\cite{radam} with an initial learning rate 1e-4.
Then, the learning rate was decreased by one-tenth at 200 and 280 epochs. 
We employed the hyperparameters $\lambda_{\text{s}}=10, \lambda_{\text{r}}=1, \lambda_{\text{ov}}=1$ and $\lambda_{\text{np}}=1$ for all experiments. 
%
For the inference $\text{PR}^3$, we used $p_1=0.4$ and $p_2=0.4$ for the random-replacing process in Figure~\ref{fig:fig4}. 
For the setting of probabilities, details are included in our supplementary material. 
The batch size and input size are 2 and $500\times500$, respectively, with random cropping, rotation, and mirroring augmentations. 
For a structure of the noise extractor of I2V, NAFNet~\cite{nafnet} is adopted for the function $h$ with a single dropout layer~\cite{reflash} before the last layer of NAFNet. $\text{I2V}^{\textbf{\text{B}}}$ denotes the proposed method with the baseline inference in experiments.

\noindent \textbf{Real-world datasets.} To validate I2V in real-world noisy image (only for sRGB) datasets, we constructed two validation scenarios: 

1) \textbf{External validation.} We employ SIDD-Medium~\cite{SIDD}, which consists of 320 noisy-clean pairs for training. 
The SIDD-validation dataset contains 1280 patches of size $256\times256$ to find proper hyper-parameters for all experiments including I2V. 
After training with proper hyperparameters, we upload the results to each site for SIDD benchmark and DND benchmark datasets~\cite{dnd}, which include 1280 noisy patches ($256\times256$) and 50 real-world noisy images, respectively.

2) \textbf{Fully self-supervised denoising.} This experiment was designed for the practical application of the proposed method. 
Without clean supervision to target noisy images, we 
compare the performance between state-of-the-art methods and I2V. 
We employed the NIND~\cite{nind} dataset which consists of 22, 14, 13, and 79 clean-noisy pairs for ISO levels of 3200, 4000, 5000, and 6400, except for 16-bit images, respectively. 
We randomly extracted 20 patches of size $512\times512$
in the test phase for 
efficient computation.

\noindent \textbf{Image quality assessment metrics.} Most denoising studies actively employ PSNR and SSIM~\cite{ssim} to measure denoising quality. Although PSNR and SSIM have been verified to measure denoising quality in the past, it is very limited to capture perceptually relevant textures because these metrics depend on pixel-wise image differences. To measure the detailed texture restoration performance, we employed LPIPS~\cite{lpips} and DISTS~\cite{dists} as deep feature-based texture and detail structure similarity metrics. For LPIPS, we set the network type to a VGG network structure~\cite{vgg}.

\begin{figure*}[t]
\centering
\includegraphics[width=17cm,keepaspectratio]{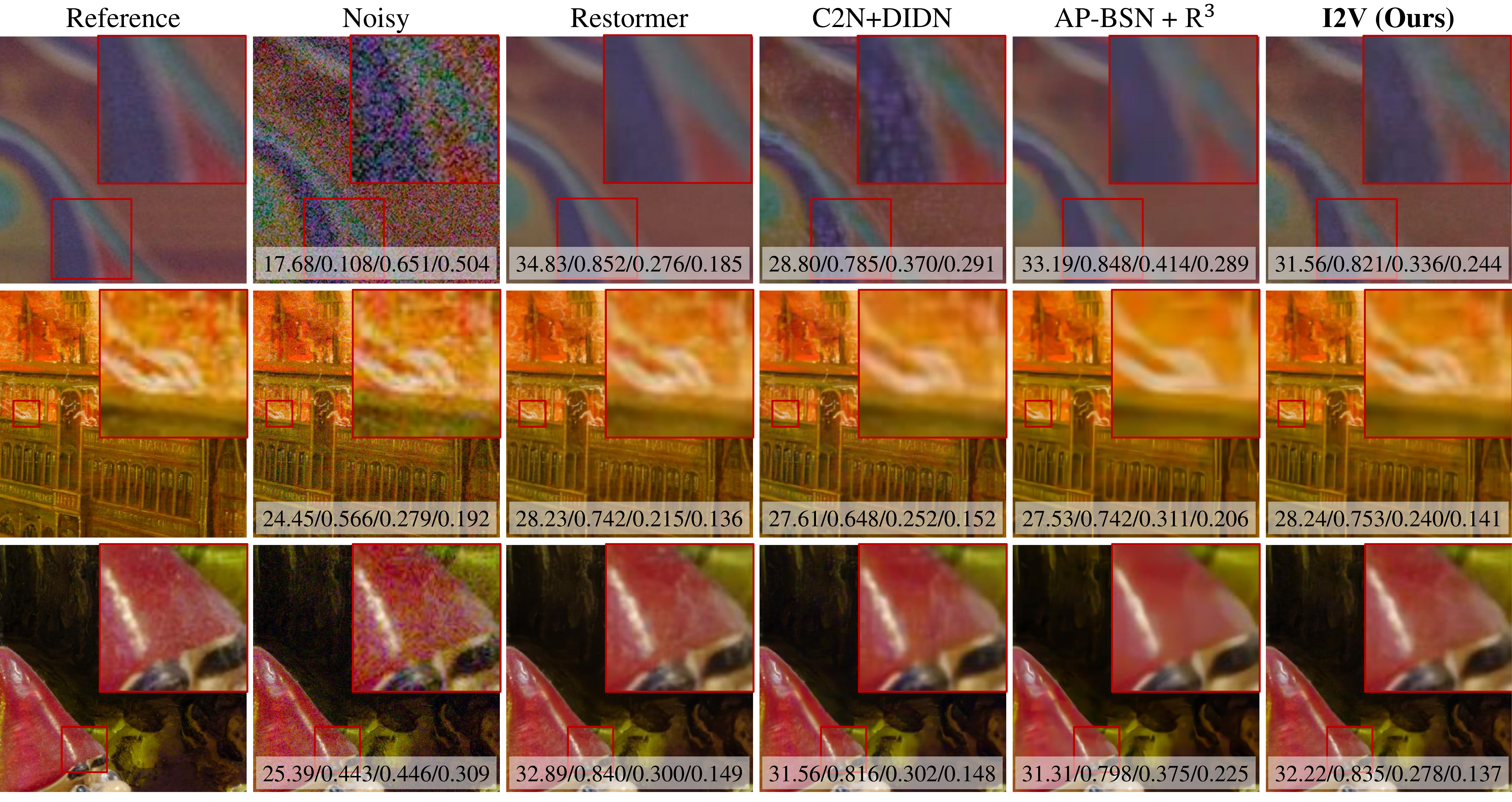}
\caption{Qualitative results for visual quality assessment. The figure of each row was chosen on SIDD validation, NIND ISO5000, and NIND ISO6400 from the first to third rows. From left to right: PSNR $\uparrow$ / SSIM $\uparrow$ / LPIPS $\downarrow$ / DISTS $\downarrow$.}
   \label{fig:fig5}
\end{figure*}

\subsection{External Validation}
In this section, we analyze real-world sRGB image denoising scenarios with supervised, unpaired, and self-supervised denoising approaches. 
All experimental results were generated 
by ourselves in the same training scheme using the author's public code except C2N which provided the pretrained model with the same external validation setting in SIDD-Medium. 
%
As shown in Table~\ref{tab:tab1}, we observe that self-supervised denoising methods without PD (N2V and Nei2Nei) fail to eliminate real-world noise because of spatial correlation. 
AP-BSN shows a higher PSNR compared to DnCNN and C2N even though DnCNN and C2N leverage clean supervision in the SIDD validation. 
However, the LPIPS and DISTS of AP-BSN are worse than those of DnCNN and C2N. 
Interestingly, removing visual artifacts by $\text{R}^3$ leads to better PSNR and SSIM than vanilla AP-BSN; however, LPIPS and DISTS are degenerated.
%
In other words, the performance of AP-BSN or the post-processing $\text{R}^3$ is insufficient to reconstruct texture details. 
%
%
In the first row of Figure~\ref{fig:fig5}, AP-BSN + $\text{R}^3$ shows over-smoothed results with higher PSNR compared to ours.
On the contrary, I2V demonstrates better LPIPS and DISTS compared to AP-BSN and several supervised learning methods, such as DnCNN, DnCNN$^\dagger$, and DANet.
Moreover, the performance of $\text{I2V}^{\textbf{\text{B}}}$ is close to that of NAFNet in terms of the perceptual similarity metrics. 
With the $\text{PR}^3$ inference scheme, I2V not only leads to the best LPIPS and DISTS in self-supervised denoising methods, but these are in second place among supervised learning approaches. 
In addition to the SIDD validation dataset, $\text{I2V}^{\textbf{\text{B}}}$ and I2V are the best or second-best in terms of PSNR and SSIM in the SIDD benchmark and DND benchmark datasets. 
More examples of the benchmark datasets are included in the supplementary material.

\begin{table}
  \centering
  \renewcommand{\arraystretch}{1.2}
  \renewcommand{\tabcolsep}{4.3pt}
  \begin{tabular}{c|c|c|c||c}
  \hline
Ablation study (w/o) & $\mathcal{L}_{r}$ & $\mathcal{L}_{ov}$ & $\mathcal{L}_{np}$ & $\text{I2V}^{\textbf{\text{B}}}$ \\ \hline
PSNR  $\uparrow$     & 29.10 & 36.12 & \underline{\textbf{36.77}} & \underline{36.63} \\ 
SSIM  $\uparrow$     & 0.648 & \underline{0.877} & \underline{0.877} & \underline{\textbf{0.888}} \\
LPIPS $\downarrow$   & 0.464 & \underline{\textbf{0.244}} & 0.295 & \underline{0.251} \\
DISTS $\downarrow$   & 0.306 & \underline{\textbf{0.211}} & 0.259 & \underline{0.218} \\ \hline
  \end{tabular}
  \caption{Ablation study of loss functions on the SIDD validation dataset. Note that each column shows the results without the corresponding loss function. The best and second best are underlined, and the best results are marked in bold.}
  \label{tab:tab2}
\end{table}
\begin{table*}
  \centering
  \renewcommand{\arraystretch}{1.2}
  \renewcommand{\tabcolsep}{3pt}
  \begin{tabular}{c|c|c|c|c|c|c|c|c|c}
    \hline
    \multirow{2}{*}{Learning Type} & \multirow{2}{*}{Methods} & \multicolumn{4}{c|}{$\text{NIND}_{\text{ISO5000}}$} & \multicolumn{4}{c}{$\text{NIND}_{\text{ISO6400}}$} \\ 
                                   &                          & PSNR $\uparrow$ & SSIM $\uparrow$ & LPIPS $\downarrow$ & DISTS $\downarrow$ & PSNR $\uparrow$ & SSIM $\uparrow$ & LPIPS $\downarrow$ & DISTS $\downarrow$   \\ \hline
    \multirow{4}{*}{Supervised}  & DnCNN     & 32.31 & 0.806  & 0.269 & 0.151 & 32.16  & 0.809  & 0.278 & 0.165 \\ 
                                 & DnCNN$^\dagger$ & 32.49 & 0.835 & 0.324 & 0.209 & 32.52 & 0.839 & 0.313 & 0.201 \\
                                 & DANet & 33.83 & 0.857 & 0.322 & 0.216 & 33.95 & 0.865 & 0.296 & 0.207 \\
                                 & NAFNet & 34.12  & 0.864 & 0.287 & 0.184 & 34.19 & 0.872 & 0.264 & 0.177 \\
                                 & Restormer & 34.01  & 0.860  & 0.290  & 0.182 & 34.12 & 0.868 & 0.267 & 0.174 \\
                                 \hline 
    \multirow{1}{*}{Unpaired image-based} 
                                  & C2N+DIDN & 33.42 & 0.846 & 0.296 & 0.184 & 33.27 & 0.857 & 0.276 & 0.177 \\ 
                                  \hline
    \multirow{6}{*}{Self-supervised}& N2V & 27.04 & 0.658 & 0.376 & 0.217 & 
                                 27.12 & 0.664 & 0.379 & 0.227 \\ 
                                 & Nei2Nei & 28.20 & 0.698 & 0.360 & 0.221 & 28.34 & 0.706 & 0.363 & 0.220\\ 
                                 & AP-BSN & 33.08 & 0.829 & 0.304 & 0.197  & 32.96 & 0.825 & 0.297 & 0.205          \\ 
                                 & AP-BSN + $\text{R}^3$ &  33.49 & 0.847 & 0.348 & 0.243   & \underline{33.56} & \underline{0.850} & 0.340 & 0.249          \\ 
                                 & $\text{I2V}^{\textbf{\text{B}}}$ (Ours)            &  \underline{33.62} & \underline{0.848} & \underline{0.287} & \underline{0.174}   & 33.46 & 0.848 & \underline{0.285} & \underline{0.184}               \\ 
                                 & I2V (Ours) & \underline{\textbf{33.74}} &\underline{\textbf{0.854}} & \underline{\textbf{0.267}} & \underline{\textbf{0.159}} & \underline{\textbf{33.62}} & \underline{\textbf{0.858}} & \underline{\textbf{0.267}} & \underline{\textbf{0.167}}  \\ \hline

  \end{tabular}
  \caption{Quantitative results on the NIND dataset. The best and second best are underlined, and the best results are marked in bold. Self-supervised denoising methods are trained by same data for testing as a fully self-supervised manner.}
  \label{tab:tab3}
\end{table*}

\begin{table}
  \centering
  \renewcommand{\arraystretch}{1.2}
  \renewcommand{\tabcolsep}{2.3pt}
  \begin{tabular}{c|c|c|c}
  \hline
Models                   & Params (M) & MACs (G) & Inf. Time (ms) \\ \hline
AP-BSN                      & 3.66   & 203.32  &  4.77            \\ 
AP-BSN + $\text{R}^3$       & 3.66   & 1829.87 &  290.45             \\
$\text{I2V}^{\textbf{\text{B}}}$ (Ours)                  & 11.03  & 207.54  &  41.88            \\ 
I2V (Ours)  & 11.03  & 219.25  &  80.09             \\ \hline
  \end{tabular}
  \caption{The multiplier-accumulator operation (MAC) and inference time (Inf. Time) are measured at each $256\times256$ patch of SIDD validation dataset.}
  \label{tab:tab4}
\end{table}

\subsection{Ablation Study}
\label{ablation_study}
We conducted ablation experiments on the SIDD validation dataset to assess the efficacy of loss functions (Table~\ref{tab:tab2}) and proposed methods, i.e., the order-variant PD process (see the supplement).
%
%
In Table~\ref{tab:tab2}, omitting $\mathcal{L}_\text{r}$ shows overall performance degradation because of the absence of detail information-based learning from BSN $f$ except keeping contents information through $\mathcal{L}_\text{ov}$.
%
%
Omitting $\mathcal{L}_\text{ov}$ shows that it improves the performance of PSNR and SSIM, whereas minimal performance degradation is observed in LPIPS and DISTS.
%
%
This is because $\mathcal{L}_\text{ov}$ promotes content similarity and leads to higher PSNR and SSIM in place of texture restoration performance improvements. 
%
Omitting $\mathcal{L}_\text{np}$ shows that the absence of the noise prior loss causes overfitting to the pseudo-noise map of BSN, making the LPIPS and DISTS results close to those of AP-BSN. 
In other words, $\mathcal{L}_\text{np}$ can successfully reduce texture deformation by preventing $h$ from learning aliasing artifacts in the pseudo-noise map.

We address the effectiveness of the order-variant $\text{PD}_s(\cdot,I)$ with a random transformation matrix $I\in\mathcal{I}$ in our supplementary material. 
%
If the order-variant PD process is replaced with the order-invariant PD ($\text{PD}_s(\cdot,I_0)$) in $\mathcal{L}_\text{s}$ and $\mathcal{L}_\text{ov}$, the overall performance is decreased. 
%
%

\subsection{Fully Self-Supervised Denoising}
For the wide adaptation of I2V, we construct the denoising experiment in a fully self-supervised manner, without an external training dataset. We suppose that only target noisy images are available in this experiment. 
%
For the supervised learning approaches and the unpaired image denoising method, we employ the SIDD-Medium dataset to train the denoisers to assume a real application scenario of the fully self-supervised manner.
Table~\ref{tab:tab3} summarizes the overall performance of the fully self-supervised denoising scenario in the NIND dataset according to ISO5000 and ISO6400. 
We provide more results for ISO3200 and ISO4000 in the supplementary material. 
In Table~\ref{tab:tab3}, I2V is ranked in first place among the state-of-the-art self-supervised denoising methods for all metrics. 
As for ISO5000, our proposed method outperforms DANet, NAFNet, and Restormer in terms of LPIPS and DISTS with slightly lower PSNR. 
In the second and third rows of Figure~\ref{fig:fig5}, the proposed I2V demonstrates texture-rich results; however, the results of AP-BSN + $\text{R}^3$ show much over-smoothed outputs compared to its reference.
Self-supervised learning-based methods employ the target noisy images directly to train deep learning models; otherwise, the supervised denoising or unpaired image denoising method is trained by clean and noisy images belonging to different datasets.
The different datasets may have different textures or structure information compared to the target noisy images. 
We believe this is the reason why the proposed method performs similarly or even better in LPIPS and DISTS compared to the supervised learning approaches.

\section{Discussion}
In the denoising performance comparisons, we demonstrate the superiority of $\text{I2V}^{\textbf{\text{B}}}$ and I2V with respect to four image quality assessment metrics. 
However, the proposed method I2V includes the additional network $h$ as well as BSN $f$, which is the same network of AP-BSN. 
This may trigger an increase in computation cost compared with AP-BSN in the inference stage. 
To measure the effectiveness of the inference cost, we investigate the number of parameters, MACs, and inference time, as shown in Table~\ref{tab:tab4}. 
The inference time was measured by RTX A6000. In Table~\ref{tab:tab4}, I2V is approximately 3.62$\times$ faster than AP-BSN+$\text{R}^3$ for denoising because we do not take multiple repetitions of the random-replacing strategy.
Furthermore, AP-BSN+$\text{R}^3$ is computationally expensive even though the number of parameters is much smaller than ours in the comparison for MACs.
As a limitation of I2V, 2.32$\times$ more GPU memory than AP-BSN is needed to train the proposed I2V in the same training setting because of the large computation cost for the proposed losses.

In Table~\ref{tab:tab1}, our reproduced results of AP-BSN and with $\text{R}^3$ are much better in the SIDD validation and SIDD benchmark data compared to the reported results in the same external validation setting. In the DND benchmark dataset, the reproduced results show worse performance because our experimental setting is external validation, while AP-BSN adopts the fully self-supervised denoising setting for the DND benchmark.

\section{Conclusion}
In this study, we verified that the proposed method I2V outperforms state-of-the-art self-supervised denoising approaches including some supervised learning methods through external validation and fully self-supervised scenarios in real-world sRGB datasets. 
Unlike current self-supervised blind denoising methods, we first compare the denoising quality using four measurement metrics to demonstrate superiority. 
Not only the visual performance, but I2V also requires a smaller amount of computational cost compared to AP-BSN which is the most recent self-supervised denoising method. 
In future work, we plan to investigate the performance using different noise extractor structures, such as Restormer. 


\section*{Acknowledgement}
This work was partially supported by the Bio \& Medical Technology Development Program of the National Research Foundation of Korea (NRF) funded by the Ministry of Science and ICT (MSIT) (NRF-2019M3E5D2A01063819, NRF-2019M3E5D2A01063794), the Basic Science Research Program through the NRF funded by the Ministry of Education (NRF-2021R1A6A1A13044830), a grant from the Korea Health Technology R\&D Project through the Korea Health Industry Development Institute (KHIDI) funded by the Ministry of Health \& Welfare (HI18C0316), the ICT Creative Consilience program (IITP-2023-2020-0-01819) of the Institute for Information \& communications Technology Planning \& Evaluation (IITP) funded by MSIT, the Korea Institute of Science and Technology (KIST) Institutional Program, Republic of Korea (2E31511), and a Korea University Grant.

\newpage
\hfill

\newpage
\renewcommand{\thetable}{S\arabic{table}}
\renewcommand{\thefigure}{S\arabic{figure}}
\setcounter{table}{0}
\setcounter{figure}{0}
\begin{figure}[t]
\includegraphics[width=8cm,keepaspectratio]{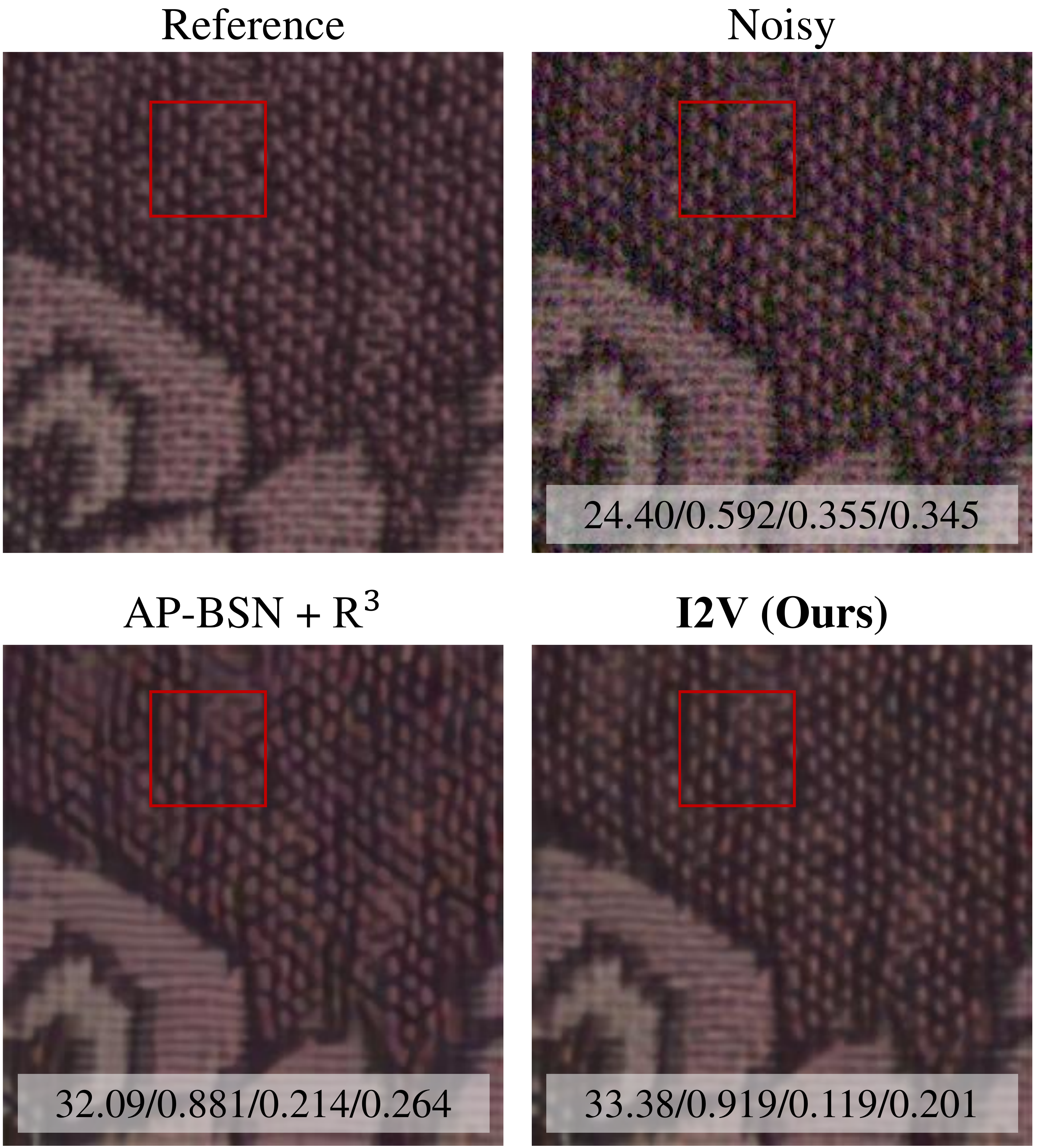}
\caption{An example of color and structure deformation shown in the result of AP-BSN + $\text{R}^3$~\cite{apbsn} in SIDD validation dataset. From left to right: PSNR (peak signal-to-noise ratio) $\uparrow$ / SSIM (structural similarity) $\uparrow$~\cite{ssim} / LPIPS (learned perceptual image patch similarity) $\downarrow$~\cite{lpips} / DISTS (deep image structure and texture similarity) $\downarrow$~\cite{dists}.}
   \label{fig:figs1}
\end{figure}

\section*{S1. Motivation of Noise Prior Loss}
\begin{figure*}[t]
    \centering
    \begin{minipage}{.245\textwidth}
        \includegraphics[width=1\textwidth]{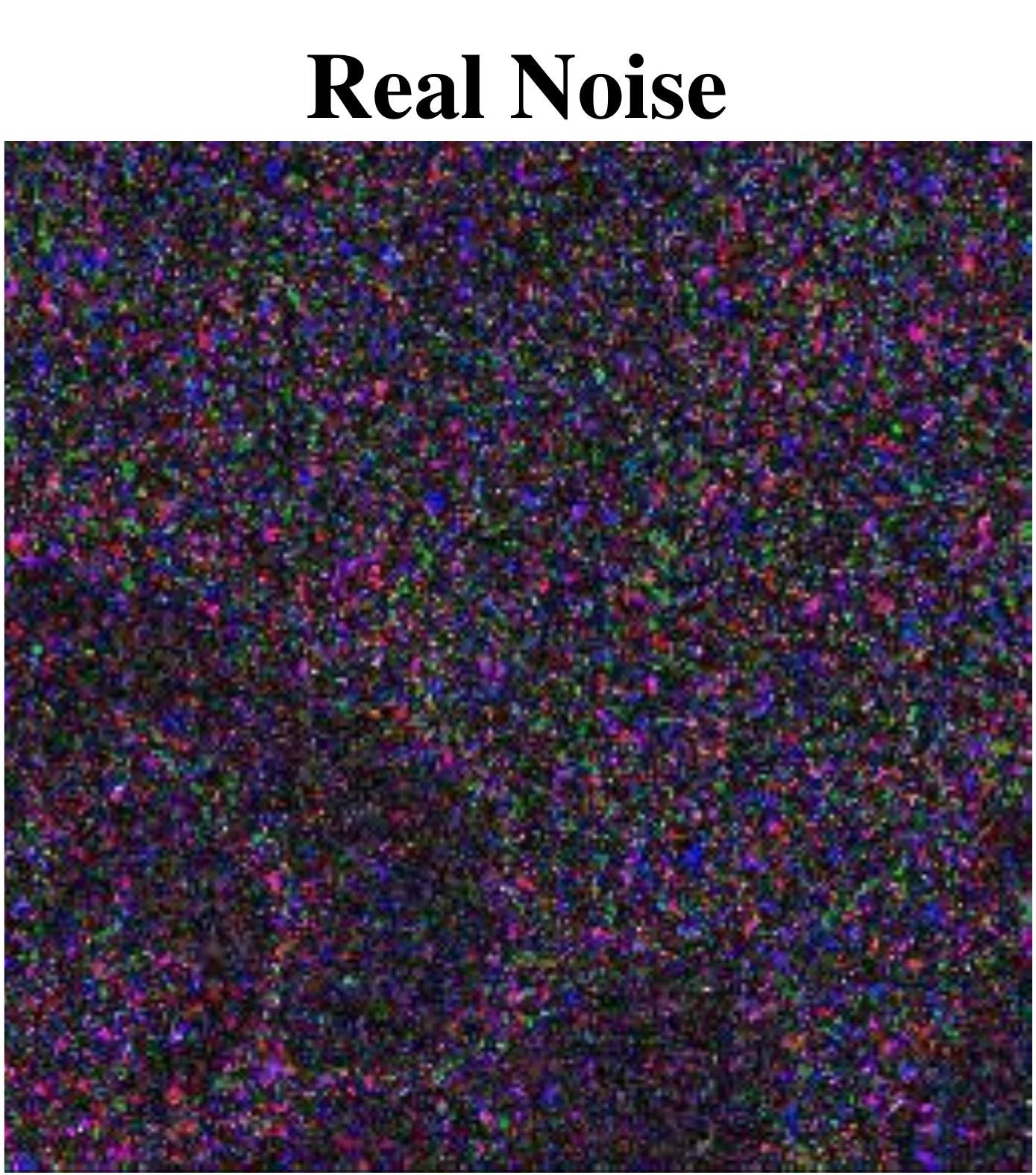}
    \end{minipage}
    \begin{minipage}{.245\textwidth}
        \includegraphics[width=1\textwidth]{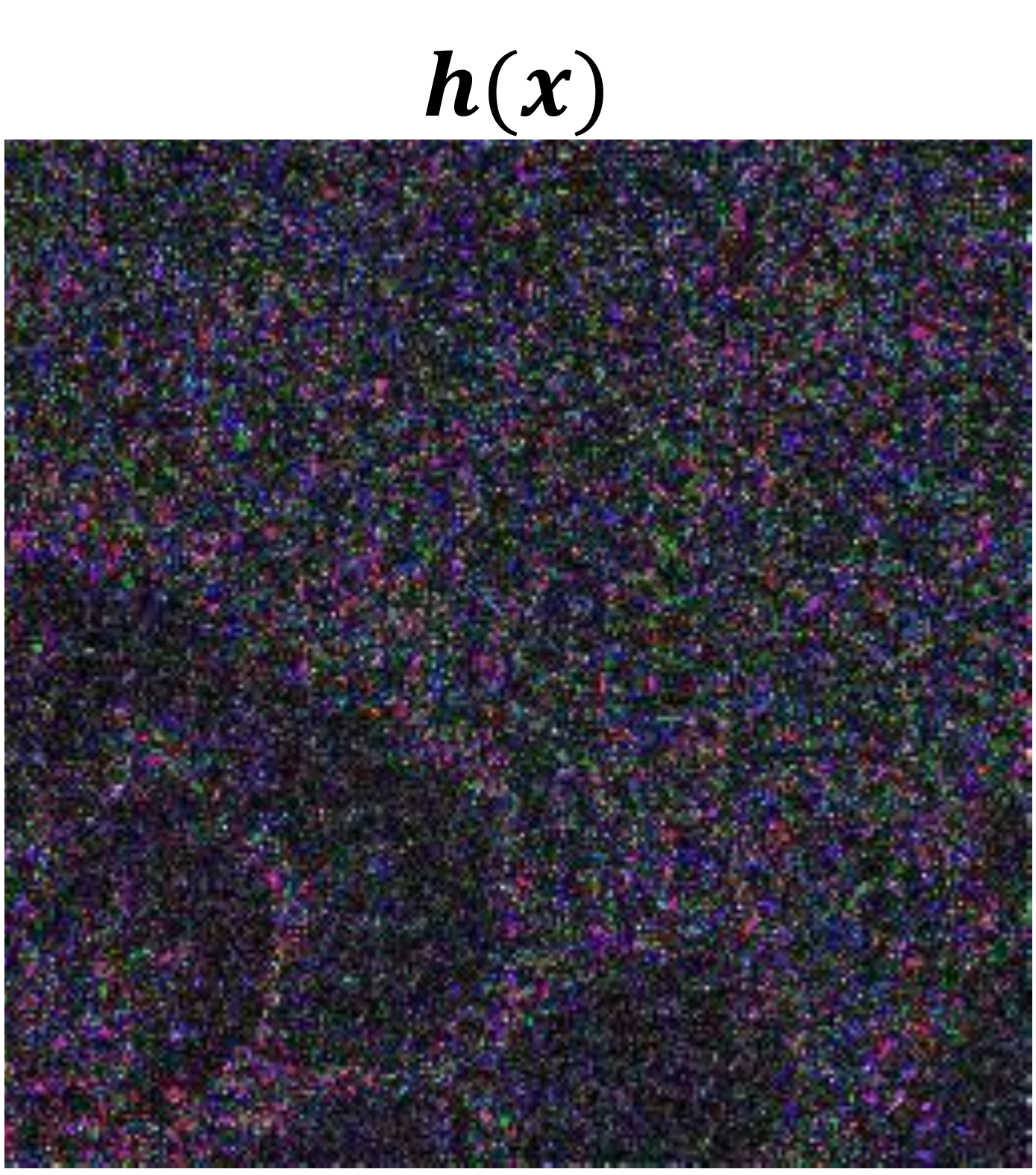}
    \end{minipage}
    \begin{minipage}{.245\textwidth}
        \includegraphics[width=1\textwidth]{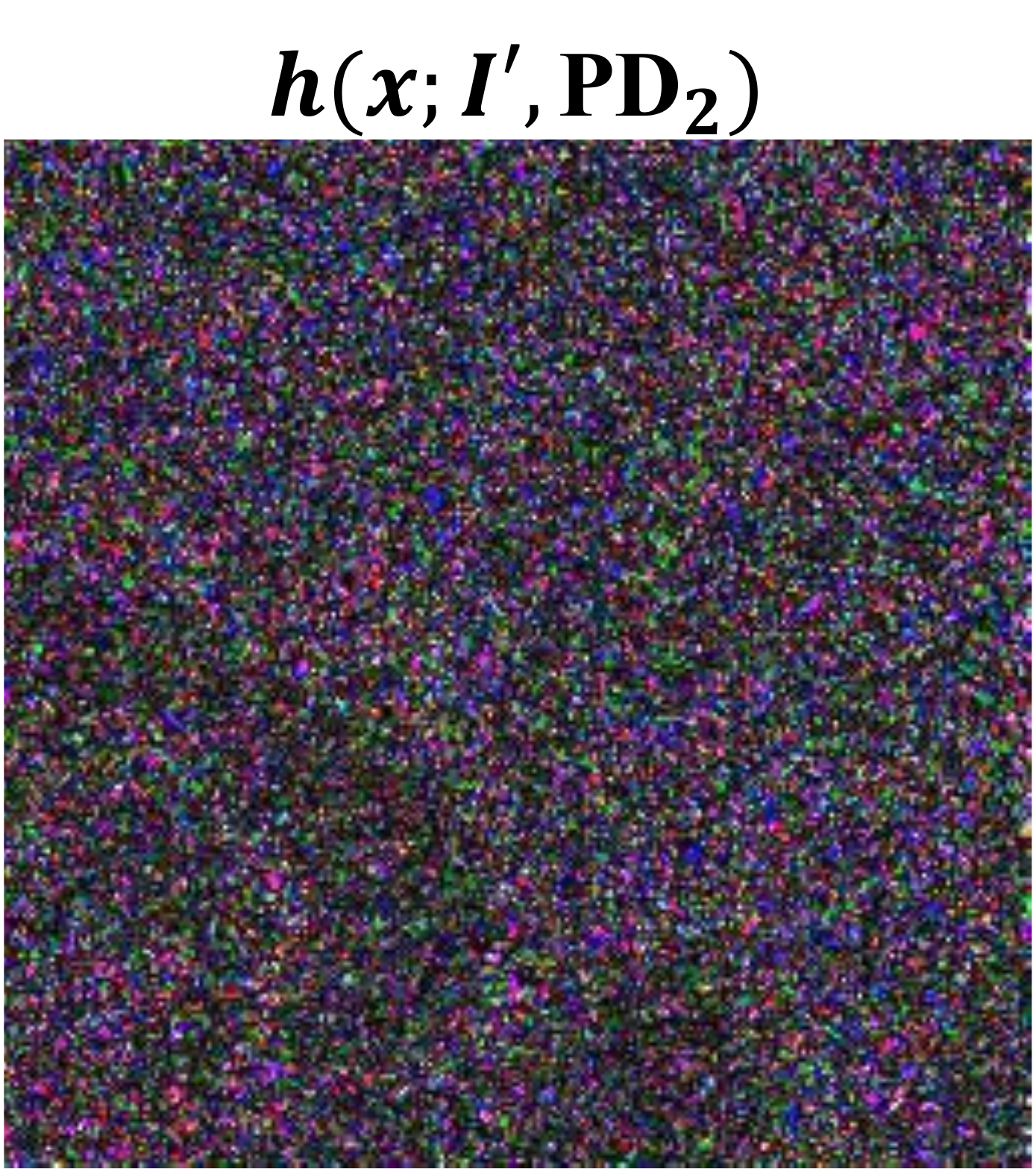}
    \end{minipage}
    \begin{minipage}{.245\textwidth}
        \includegraphics[width=1\textwidth]{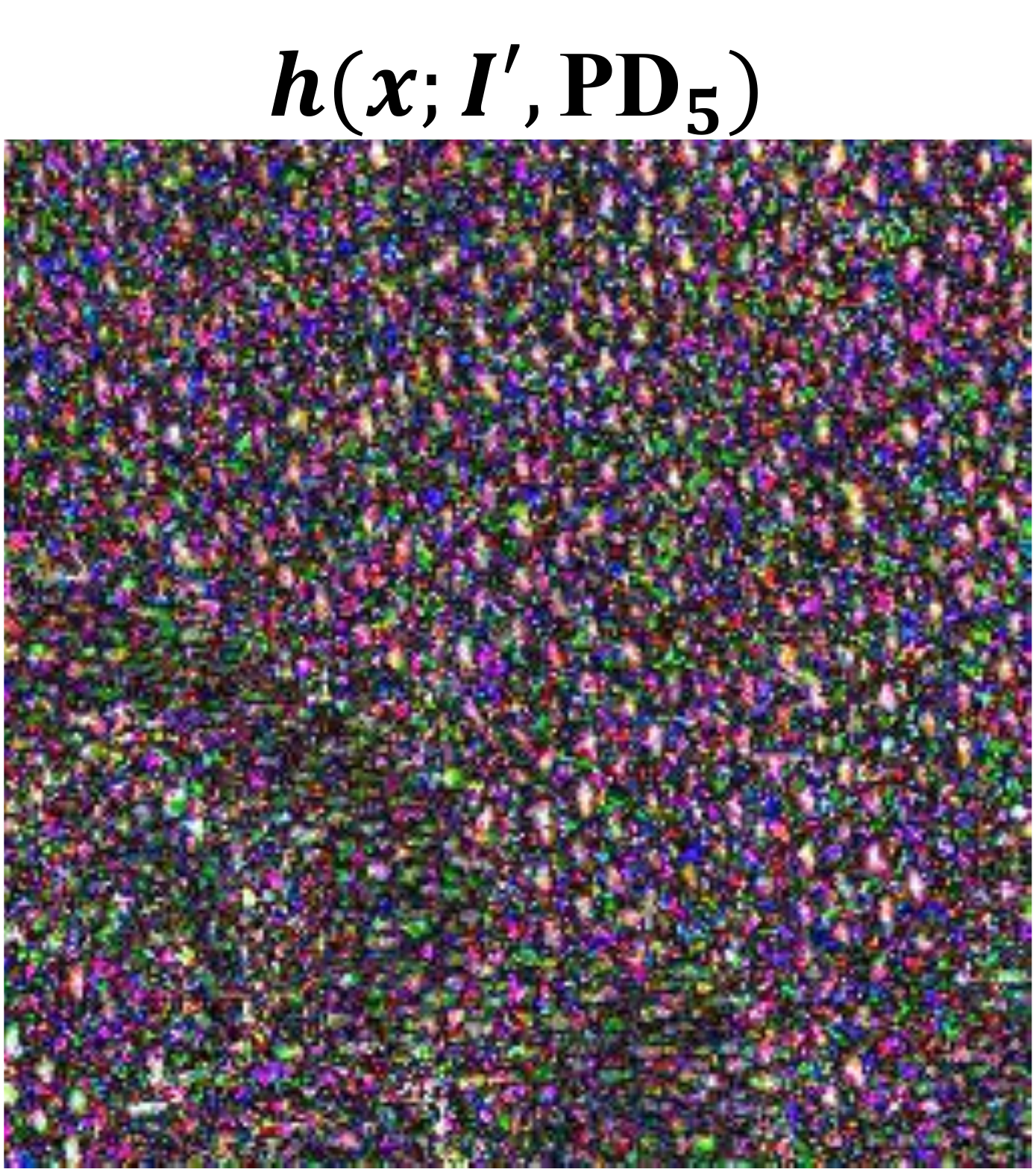}
    \end{minipage}
    \begin{minipage}{.245\textwidth}
        \includegraphics[width=1\textwidth]{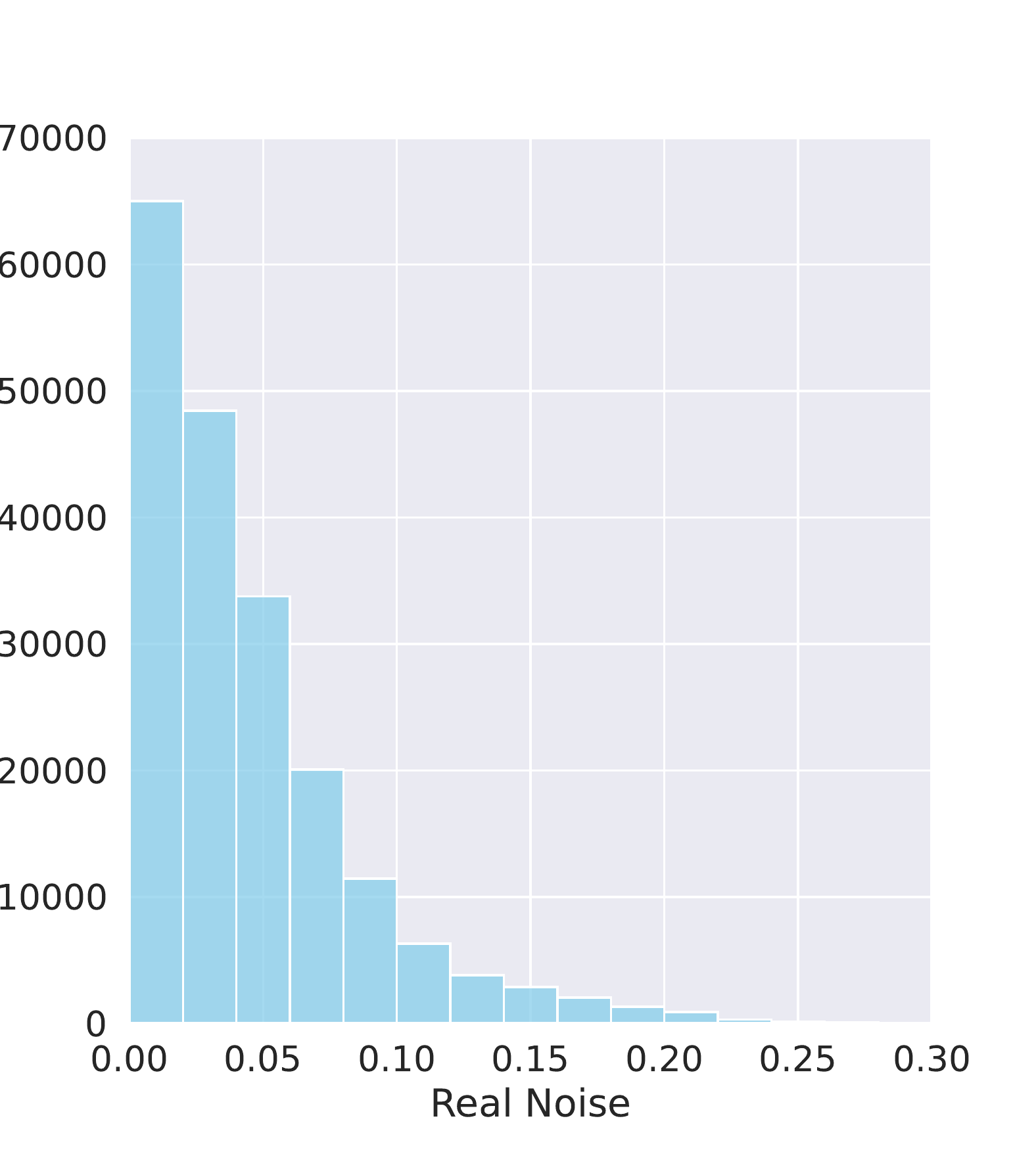}
    \end{minipage}
    \begin{minipage}{.245\textwidth}
        \includegraphics[width=1\textwidth]{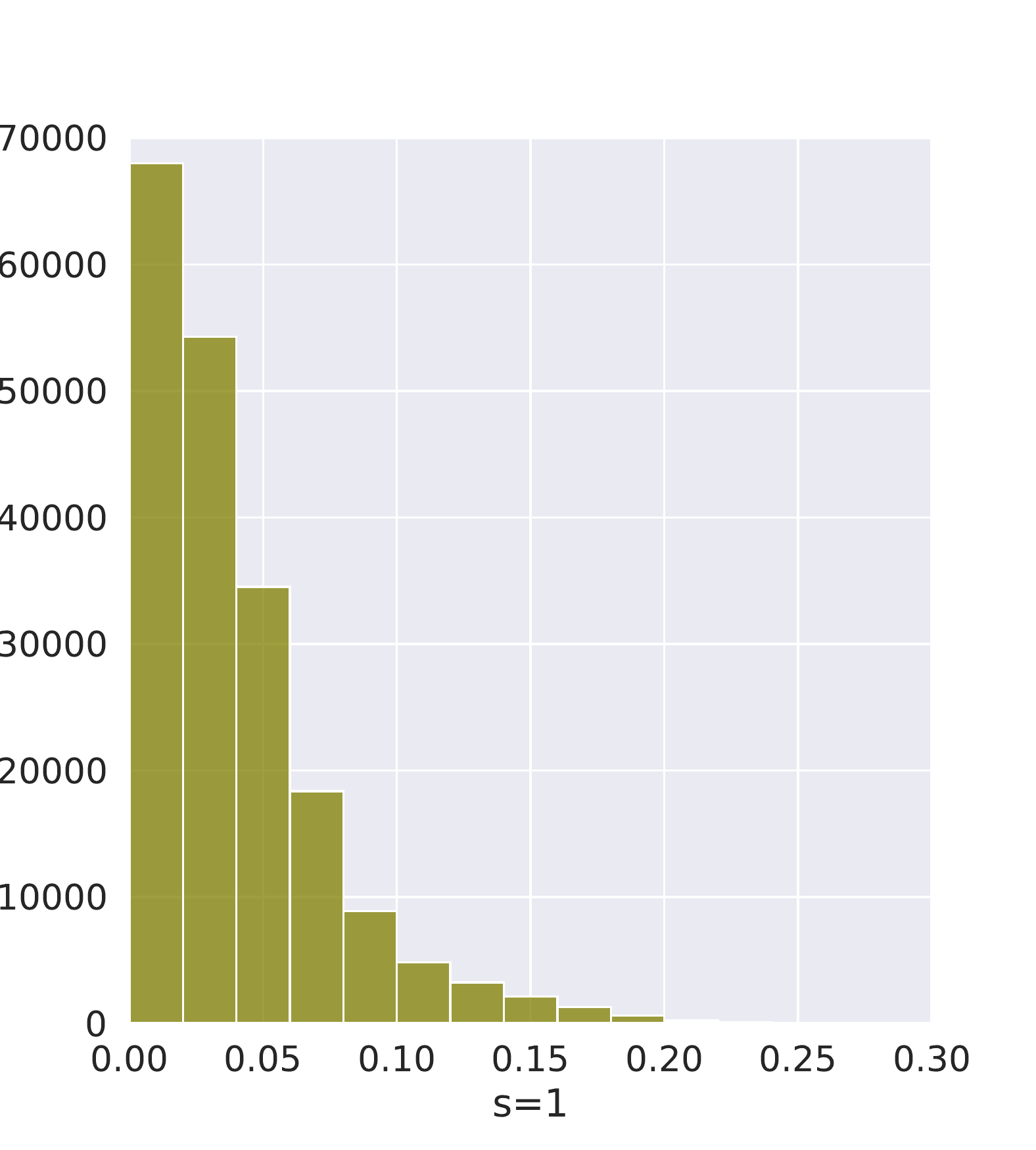}
    \end{minipage}
    \begin{minipage}{.245\textwidth}
        \includegraphics[width=1\textwidth]{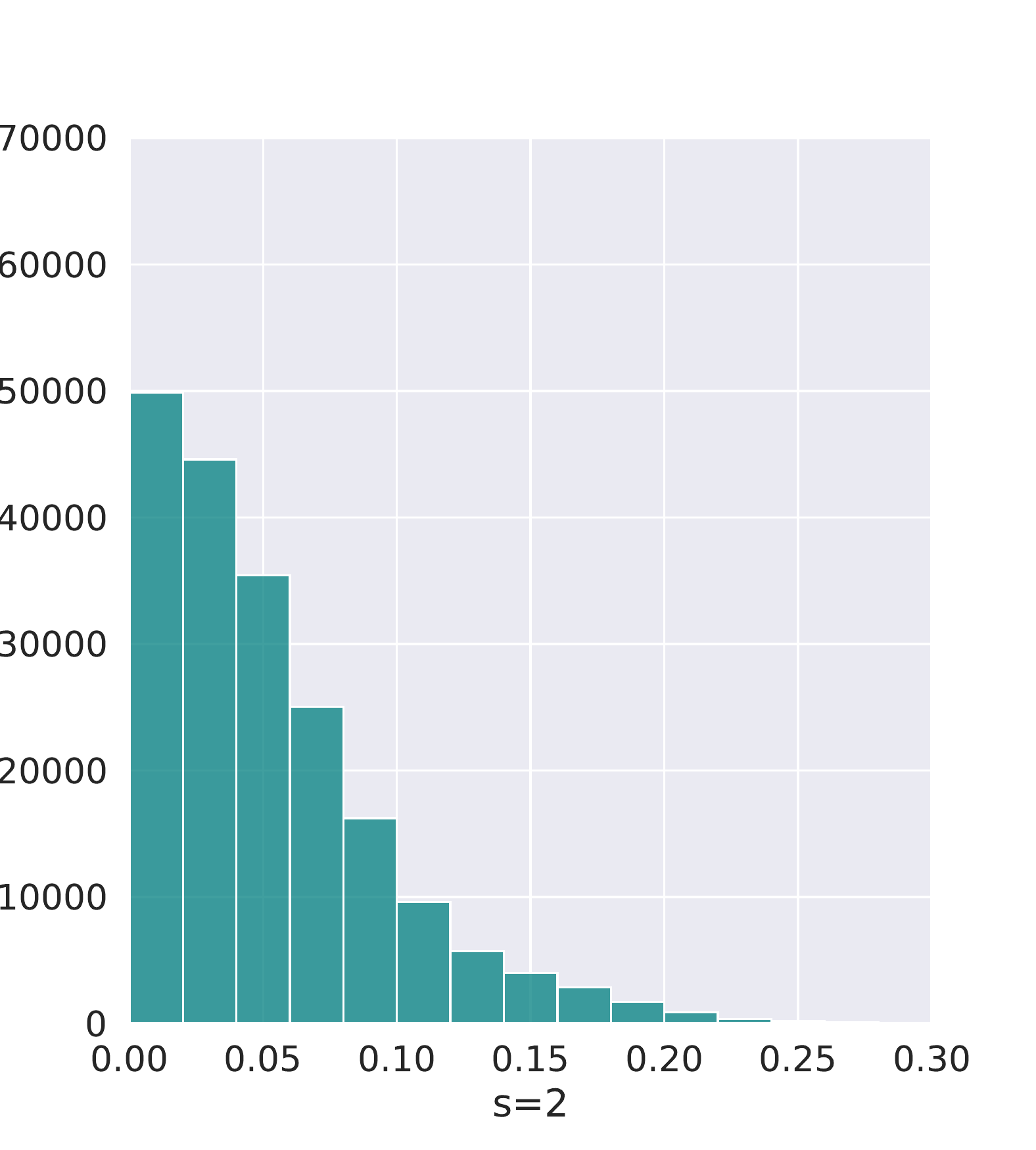}
    \end{minipage}
    \begin{minipage}{.245\textwidth}
        \includegraphics[width=1\textwidth]{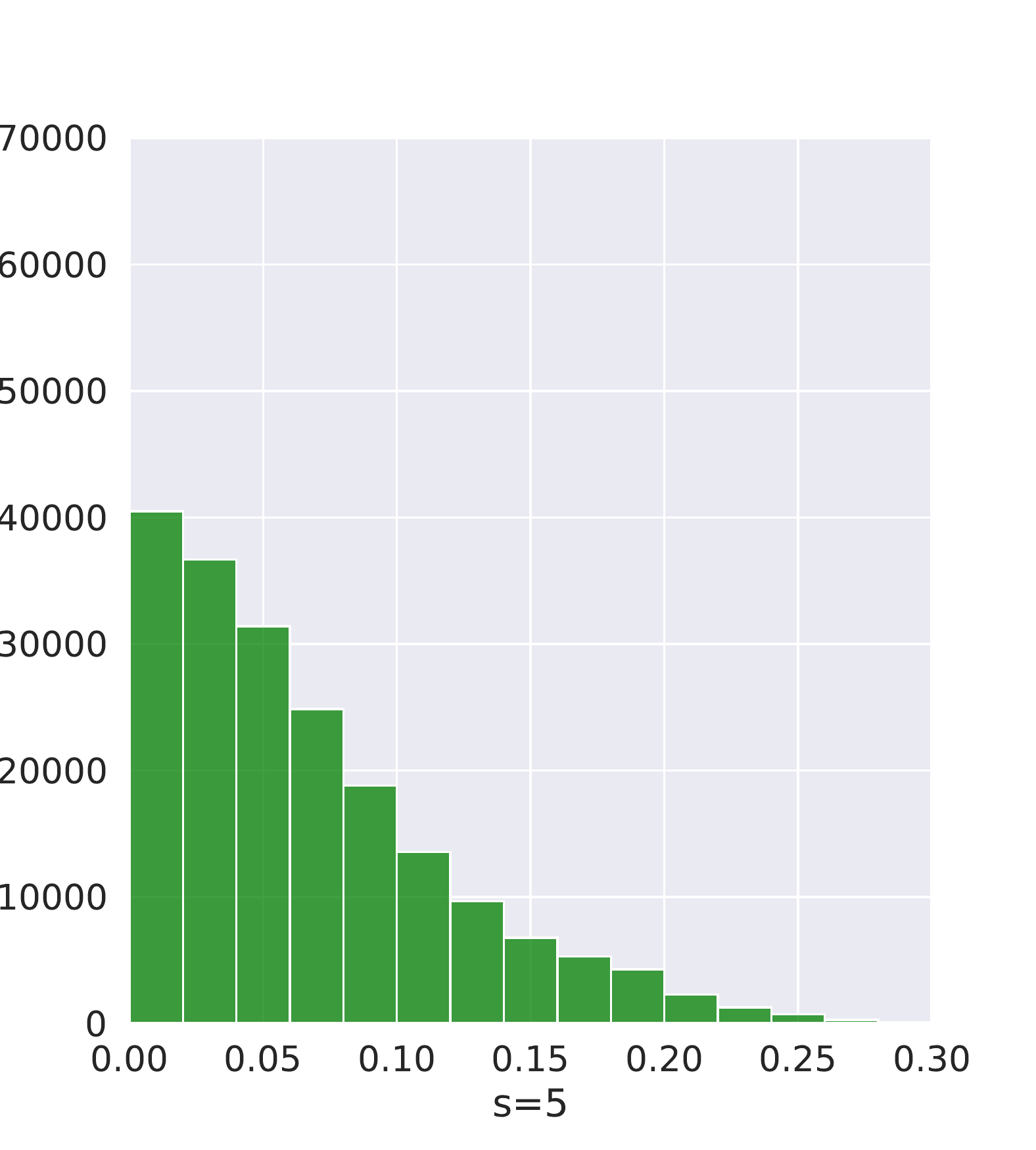}
    \end{minipage}
    \caption{Noise image and its histogram (magnitude) extracted by the noise extractor $h$ with each stride factor s in the same observation of Figure~\ref{fig:figs1}.}
    \label{fig:figs2}
\end{figure*}

In Figure~\ref{fig:figs1}, AP-BSN showed some failure cases 
to predict the correct color and details in the texture-rich sample. 
To examine the cause of the failure case, we visualize the noise maps that consist of real noise and aliasing artifacts for 
each stride factor in Figure~\ref{fig:figs2}. 
Because the aliasing artifacts are pixel-wise independent noise, a denoiser may regard the texture information as noise. 
Therefore, aliasing artifacts introduced by $\text{PD}$ change the noise distribution in the image, as shown in the histograms of Figure~\ref{fig:figs2}. 
Pseudo-noise maps generated by the PD process in the proposed self-residual learning contain the aliasing artifacts affecting the distribution of predicted noise by the noise extractor. 
Imperfect pseudo-noise labels could induce spatially different textures or colors compared to real noise. 
We discovered that the proposed noise prior loss function could limit the distribution change introduced by aliasing artifacts in self-residual learning, resulting in better LPIPS and DISTS performance as shown in Table 2 in the main text. 

\begin{table}
  \centering
  \renewcommand{\arraystretch}{1.2}
  \renewcommand{\tabcolsep}{4.3pt}
  \begin{tabular}{c|c|c|c|c}
  \hline
Ablation study       & $f(x,I_0,\text{PD}_2)$ & $x-h(x)$ & $\text{I2V}^\textbf{\text{B}}$ & I2V \\ \hline
PSNR  $\uparrow$     & 36.34 &  36.15 & \textbf{\underline{36.63}} & \underline{36.48} \\ 
SSIM  $\uparrow$     & 0.874 &  0.877 & \underline{0.888} & \textbf{\underline{0.889}} \\
LPIPS $\downarrow$   & 0.279 &  \underline{0.247} & 0.251 & \textbf{\underline{0.245}} \\
DISTS $\downarrow$   & 0.260 &  \underline{0.204} & 0.218 & \textbf{\underline{0.199}} \\ \hline
  \end{tabular}
  \caption{Results of ablation studies for each network of I2V on the SIDD validation dataset. $\text{I2V}^{\textbf{\text{B}}}$ represents I2V with the baseline inference scheme in place of $\text{PR}^3$. The best and second best are underlined, and the best results are marked in bold.}
  \label{stab:tab1}
\end{table}

\begin{table}
  \centering
  \renewcommand{\arraystretch}{1.2}
  \renewcommand{\tabcolsep}{4.3pt}
  \begin{tabular}{c|c|c}
  \hline
Ablation study       & Order-invariant PD & Order-variant PD  \\ \hline
PSNR  $\uparrow$     & 35.85 &  36.63 \\ 
SSIM  $\uparrow$     & 0.884 &  0.888 \\
LPIPS $\downarrow$   & 0.268 &  0.251 \\
DISTS $\downarrow$   & 0.241 &  0.214 \\ \hline
  \end{tabular}
  \caption{Comparison Results of using the order-invariant and order-variant PD process for $\mathcal{L}_{\text{s}}$ and $\mathcal{L}_{\text{ov}}$ on the SIDD validation dataset.}
  \label{stab:tab2}
\end{table}

\begin{figure*}[t]
\centering
    \begin{subfigure}[t]{0.45\textwidth}
        \includegraphics[width=1\textwidth]{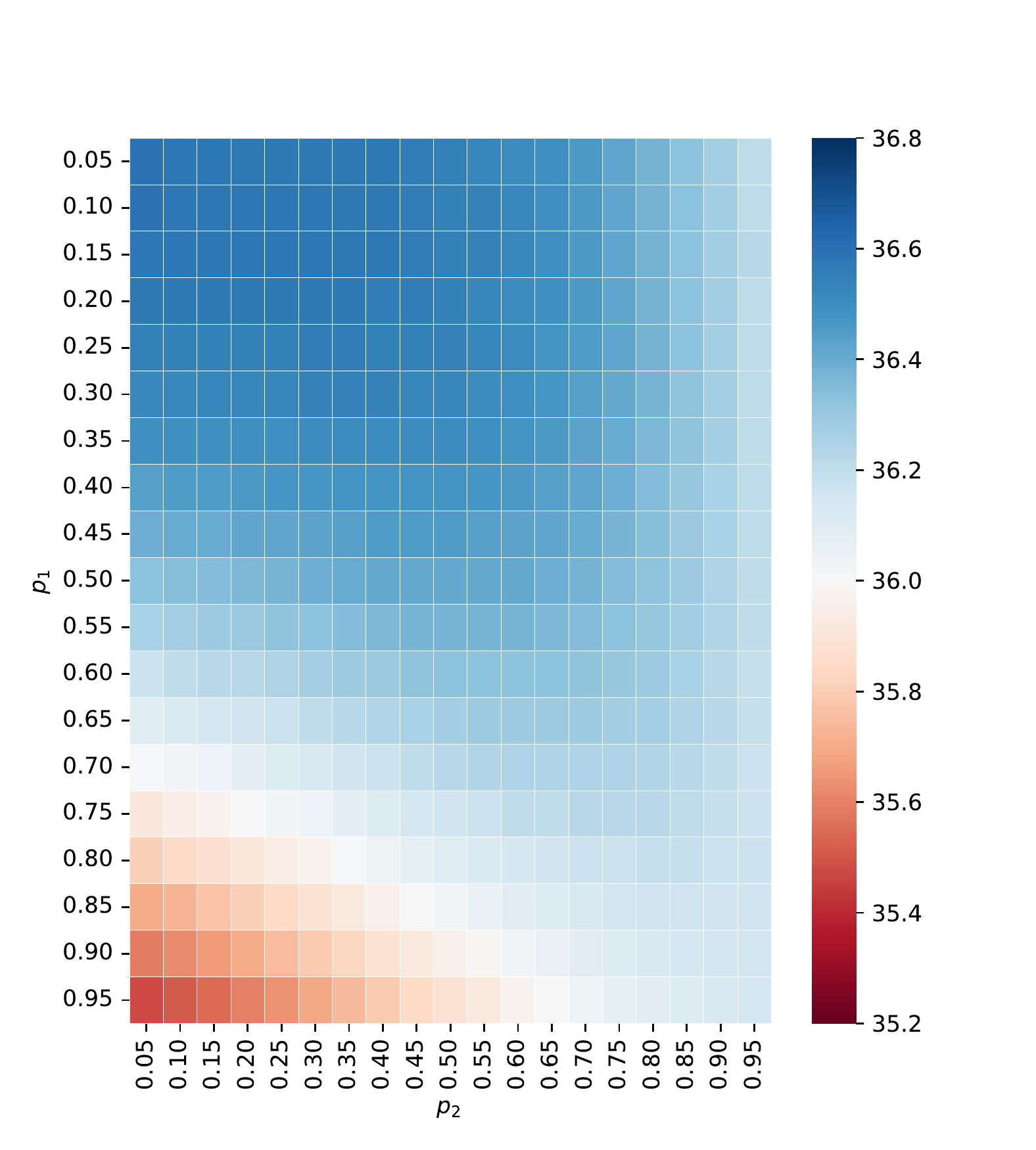}
        \caption{PSNR $\uparrow$}
    \end{subfigure}
    \begin{subfigure}[t]{0.45\textwidth}
        \includegraphics[width=1\textwidth]{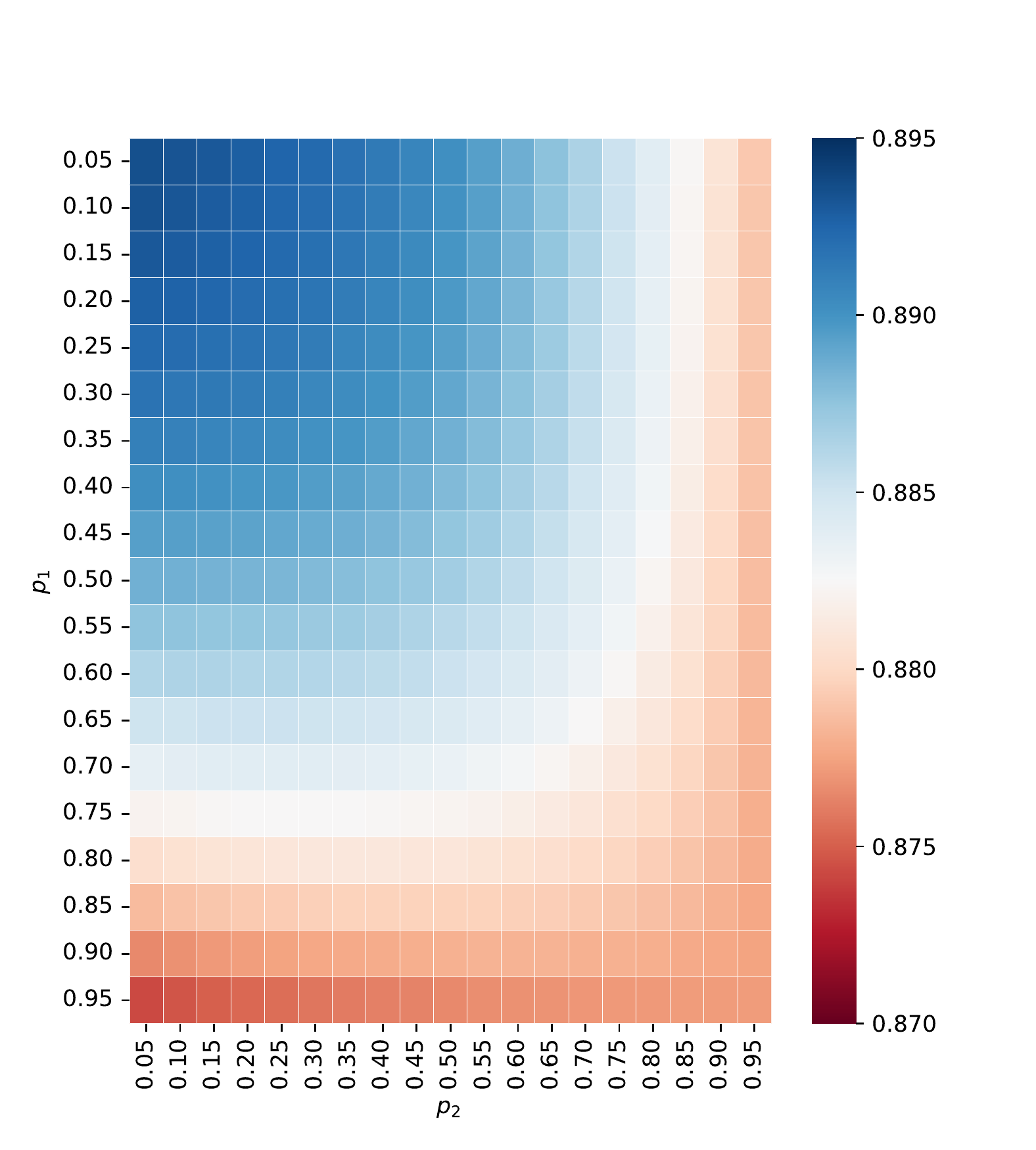}
        \caption{SSIM $\uparrow$}
    \end{subfigure}
    \begin{subfigure}[t]{0.45\textwidth}
        \includegraphics[width=1\textwidth]{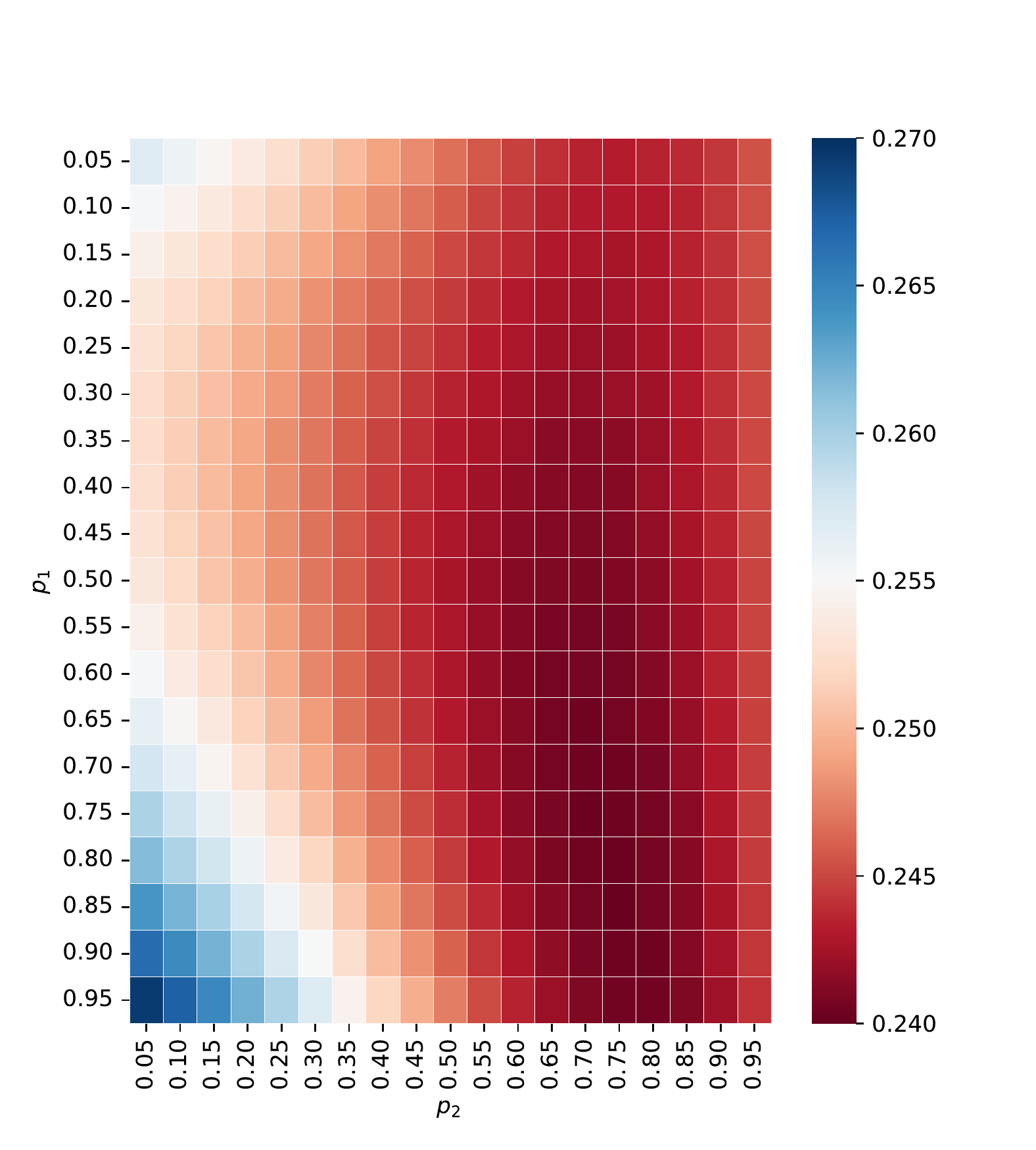}
        \caption{LPIPS $\downarrow$}
    \end{subfigure}
    \begin{subfigure}[t]{0.45\textwidth}
        \includegraphics[width=1\textwidth]{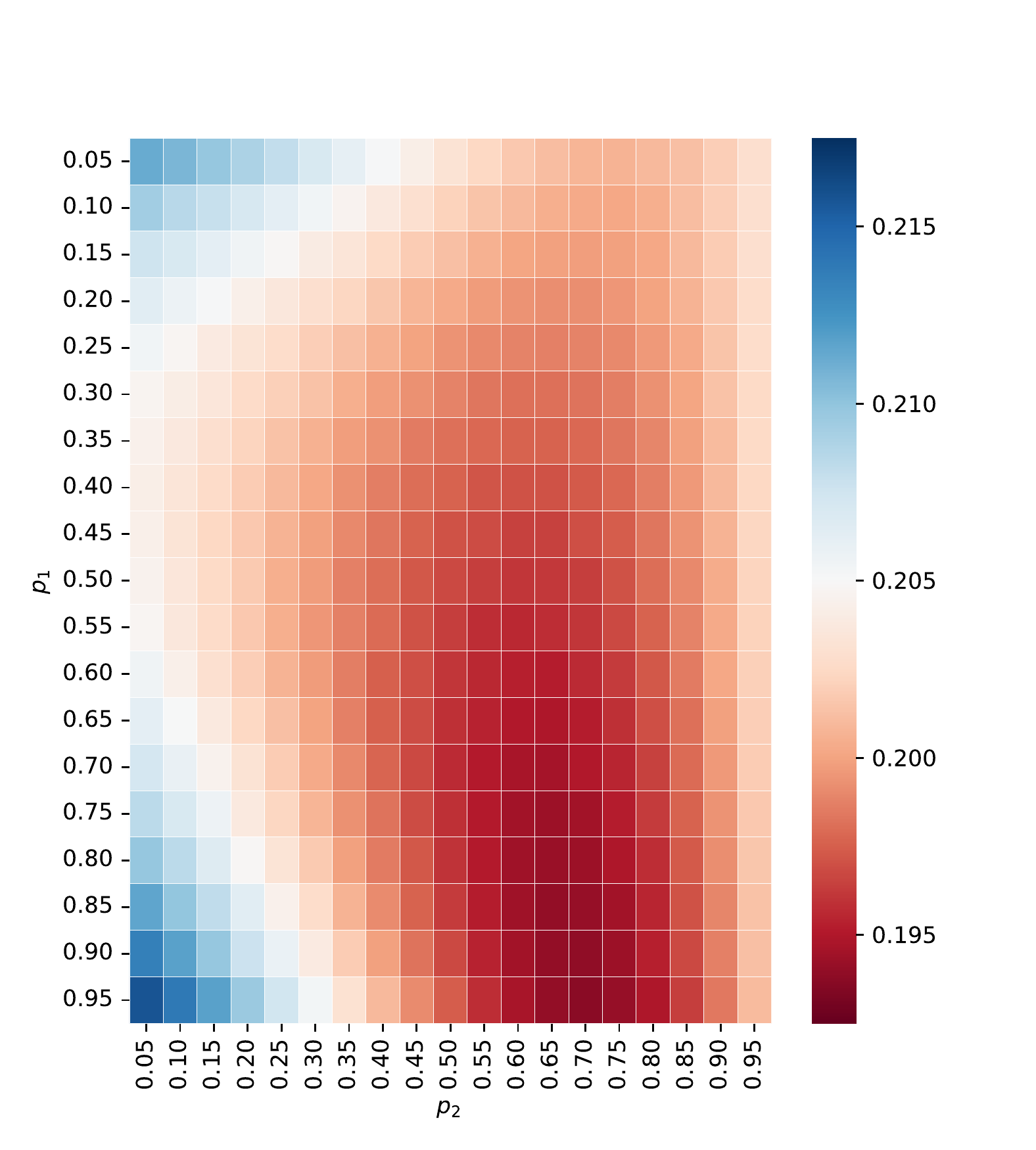}
        \caption{DISTS $\downarrow$}
    \end{subfigure}
    \caption{Heat map visualization of each error metric with respect to the choice of probabilities $p_1$ and $p_2$ for the proposed inference scheme $\text{PR}^3$ in the SIDD validation dataset.}
    \label{fig:figs3}
\end{figure*}

\section*{S2. Ablation Study}
Here we provide additional results of the ablation study in the SIDD validation dataset. Table~\ref{stab:tab1} summarizes the performance of BSN $f$, the noise extractor $h$, $\text{I2V}^{\textbf{\text{B}}}$, and I2V. The noise extractor achieves better SSIM, LPIPS, and DISTS scores than BSN $f$. The simple blending result between $f$ and $h$ (i.e., $\text{I2V}^{\textbf{\text{B}}}$) shows better PSNR and SSIM than each performance of $f$ and $h$, furthermore, LPIPS and DISTS are similar to the performance of the noise extractor. The order-variant PD process in $\mathcal{L}_{\text{s}}$ and $\mathcal{L}_{\text{ov}}$ shows performance improvements in terms of PSNR, SSIM, LPIPS, and DISTS, as shown in Table~\ref{stab:tab2}. 

\begin{table*}
  \centering
  \renewcommand{\arraystretch}{1.4}
  \renewcommand{\tabcolsep}{4.3pt}
  \begin{tabular}{c|c|c|c|c|c|c|c|c|c}
    \hline
    \multirow{2}{*}{Learning Type} & \multirow{2}{*}{Methods} & \multicolumn{4}{c|}{$\text{NIND}_{\text{ISO3200}}$} & \multicolumn{4}{c}{$\text{NIND}_{\text{ISO4000}}$} \\ 
                                   &                          & PSNR $\uparrow$ & SSIM $\uparrow$ & LPIPS $\downarrow$ & DISTS $\downarrow$ & PSNR $\uparrow$ & SSIM $\uparrow$ & LPIPS $\downarrow$ & DISTS $\downarrow$   \\ \hline
    \multirow{5}{*}{Supervised}  & DnCNN~\cite{zhang2017beyond}  &  33.82 & 0.858 & 0.216 & 0.131 & 33.38  & 0.845              & 0.235 & 0.137 \\ 
                                 & DnCNN$^\dagger$~\cite{zhouetal} & 33.53 & 0.849 & 0.293 & 0.195 & 32.93 & 0.843 & 0.295 & 0.195 \\
                                 & DANet~\cite{danet}  & 35.06 & 0.879 & 0.267 & 0.192 & 34.52 & 0.868 & 0.284 & 0.198 \\
                                 & NAFNet~\cite{nafnet} & 35.04  & 0.880  & 0.251  & 0.174  & 34.82 & 0.878 & 0.253 & 0.170 \\
                                 & Restormer~\cite{restormer} & 35.05  & 0.880  & 0.251  & 0.172 & 34.71 & 0.874 & 0.260 & 0.171 \\
                                 \hline 

    \multirow{1}{*}{Unpaired image-based} 
                                      & C2N+DIDN~\cite{c2n} & 34.86 & 0.875 & 0.260 & 0.174 & 33.97 & 0.866 & 0.269 & 0.171 \\ \hline
    
    \multirow{6}{*}{Self-supervised}& N2V~\cite{noise2void} & 28.42 & 0.766 & 0.318 & 0.196 & 
                                 27.80 & 0.736 & 0.346 & 0.198\\ 
                                 & Nei2Nei~\cite{neighbor2neighbor} & 29.47 & 0.770 & 0.310 & 0.190 & 29.38 & 0.753 & 0.328 & 0.189 \\ 
                                 & AP-BSN ~\cite{apbsn}                &  33.98 & 0.832 & 0.287  & 0.194 & 33.42 & 0.827  & 0.280 & 0.187       \\ 
                                 & AP-BSN + $\text{R}^3$~\cite{apbsn}  &  34.41 & 0.854 & 0.329  & 0.237 & \underline{\textbf{33.92}} & 0.847  & 0.333 & 0.236       \\ 
                                 & $\text{I2V}^{\textbf{\text{B}}}$ (Ours)             &  \underline{34.43} & \underline{0.855} & \underline{0.274}  & \underline{0.178} & 33.54 & \underline{0.849}  & \underline{0.279} & \underline{0.176}    \\ 
                                 & I2V (Ours) & \underline{\textbf{34.56}} & \underline{\textbf{0.868}}  & \underline{\textbf{0.251}} & \underline{\textbf{0.164}} & \underline{33.64}& \underline{\textbf{0.861}} & \underline{\textbf{0.257}} & \underline{\textbf{0.162}} \\ \hline
                                 
  \end{tabular}
  \caption{Quantitative results for ISO3200 and ISO4000 on the NIND dataset. The best and second best are underlined, and the best results are marked in bold. Self-supervised denoising methods are trained by same data for testing as a fully self-supervised manner. $\dagger$ indicates a trained network by synthetic noise (AWGN, random value impulse noise) with PD refinement. $\text{I2V}^{\textbf{\text{B}}}$ represents I2V with the baseline inference scheme in place of $\text{PR}^3$. }
  \label{stab:tab3}
\end{table*}

\section*{S3. Hyperparameters for $\text{PR}^3$}
The proposed $\text{PR}^3$ inference scheme requires a proper choice of probabilities (used in random replacing) for the best performance. 
Unfortunately, we discover a trade-off between conventional metrics (i.e., PSNR and SSIM) and perception-based metrics (i.e., LPIPS and DISTS) as shown in Figure~\ref{fig:figs3}. 
The first row of Figure~\ref{fig:figs3} shows that the proper combination between $p_1$ and $p_2$ is located on the left-top side for better 
PSNR and SSIM. 
However, the appropriate probabilities for LPIPS and DISTS 
are on the right-bottom side. 
Interestingly, PSNR and SSIM show a similar trend 
while LPIPS and DISTS show 
a similar trend. 
In our experiments, we focus on overall performance improvement rather than 
improvement biased to a specific metric. 
Therefore, we set the probabilities $p_1$ and $p_2$ to $0.4$ and $0.4$, respectively.

\begin{figure*}[t]
\centering
\includegraphics[width=15.5cm,keepaspectratio]{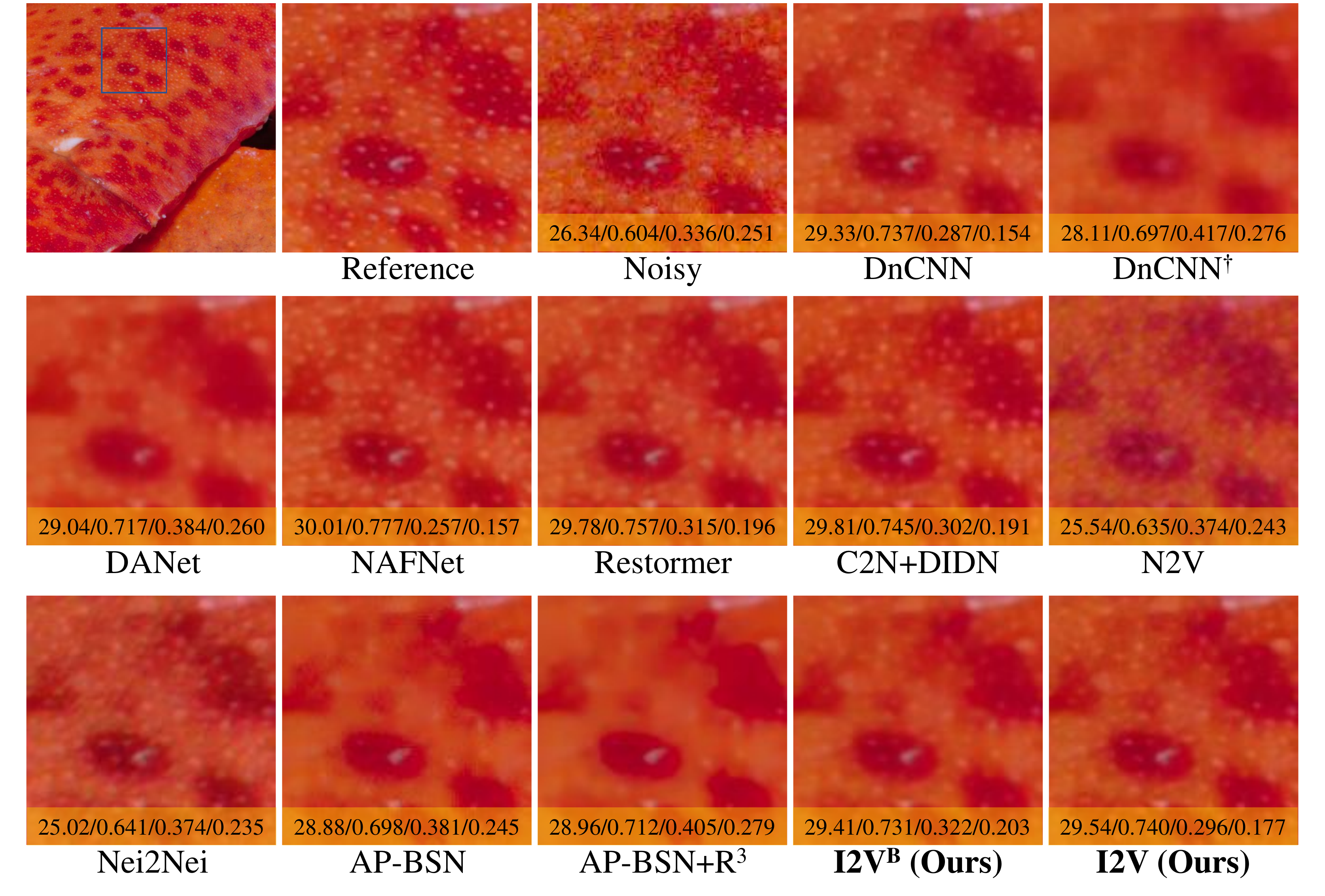}
\includegraphics[width=15.5cm,keepaspectratio]{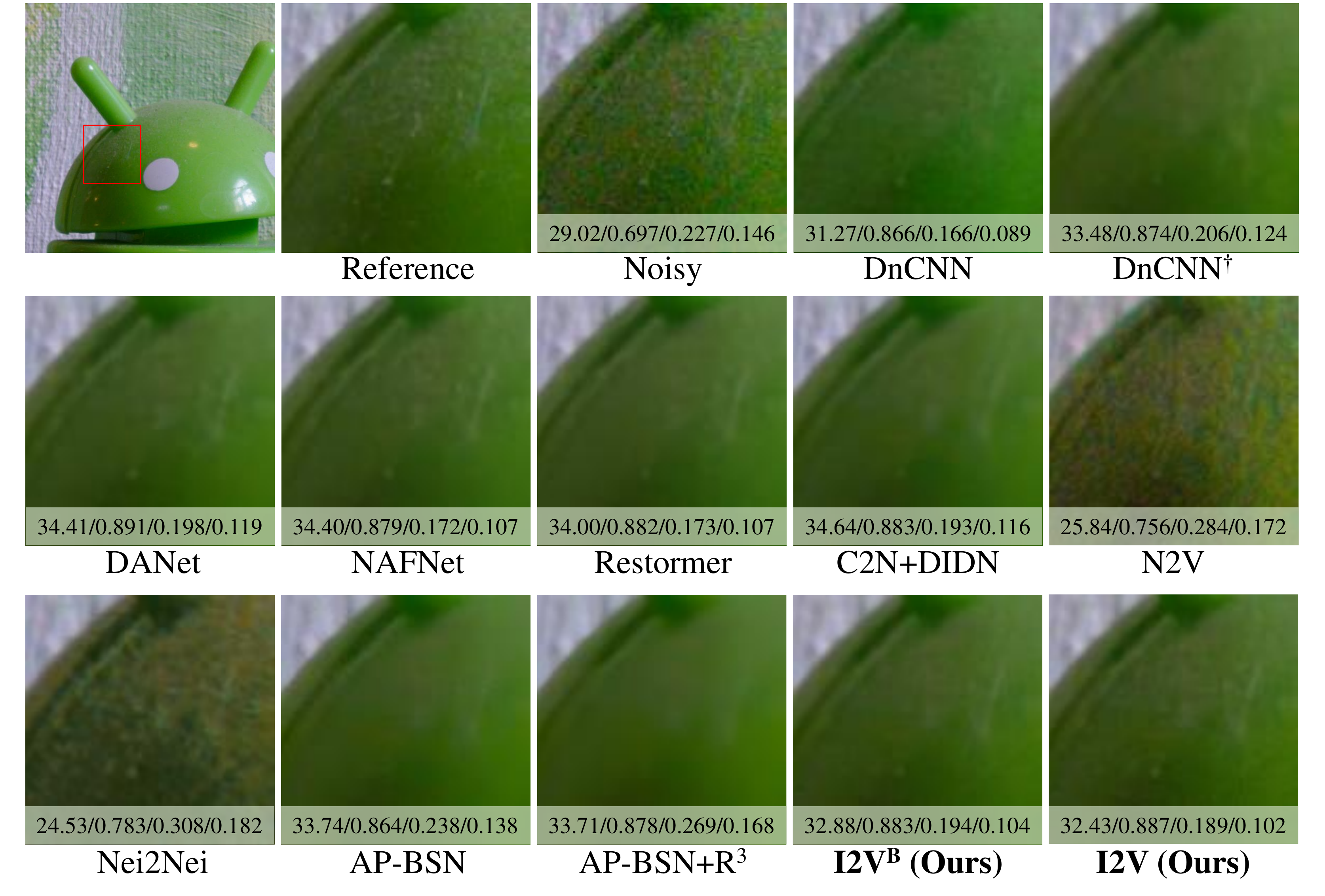}
\caption{Qualitative results for all comparison methods and our methods. Images are from the NIND ISO3200 dataset.}
\label{fig:figs4}
\end{figure*}

\begin{figure*}[t]
\centering
\includegraphics[width=15.5cm,keepaspectratio]{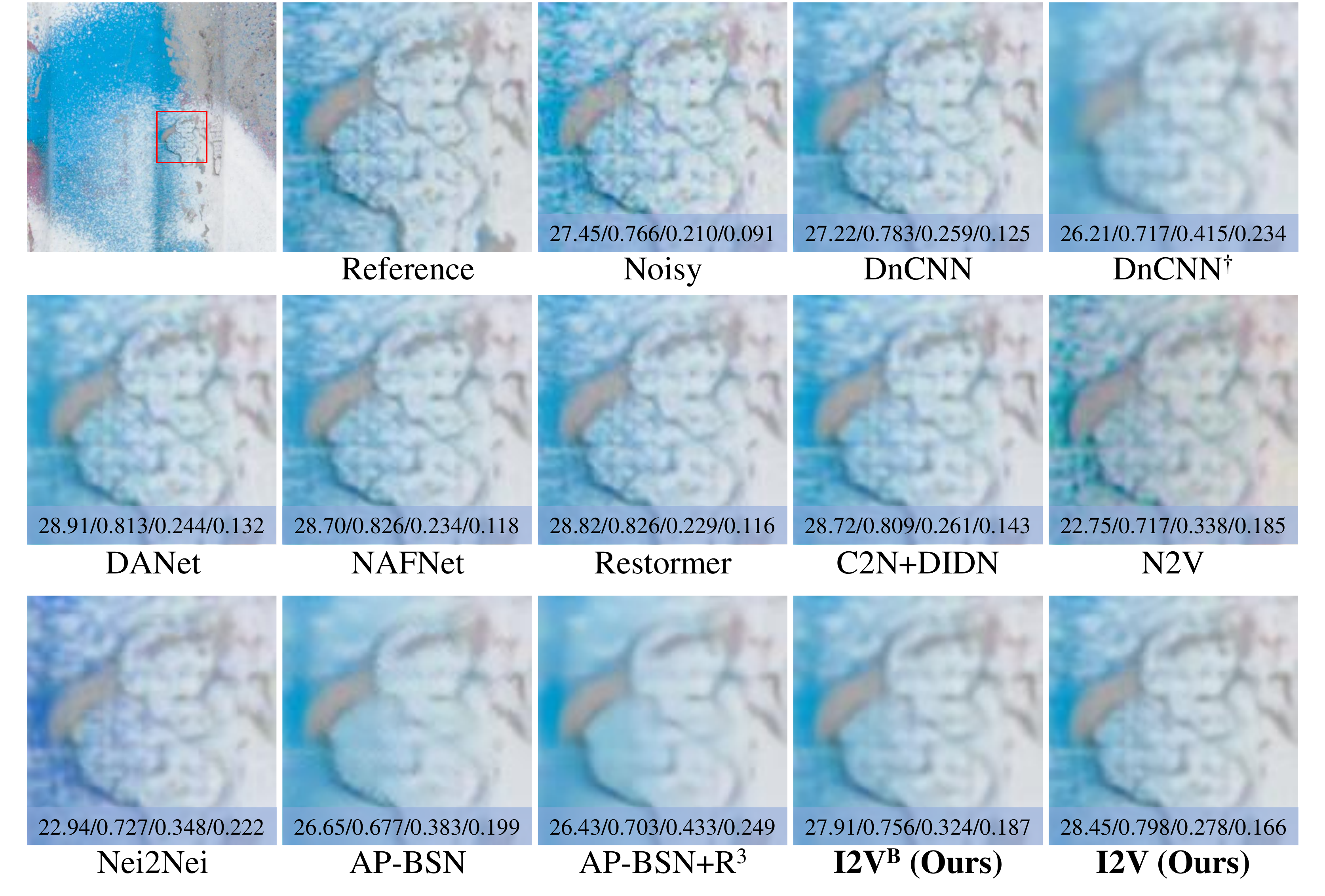}
\includegraphics[width=15.5cm,keepaspectratio]{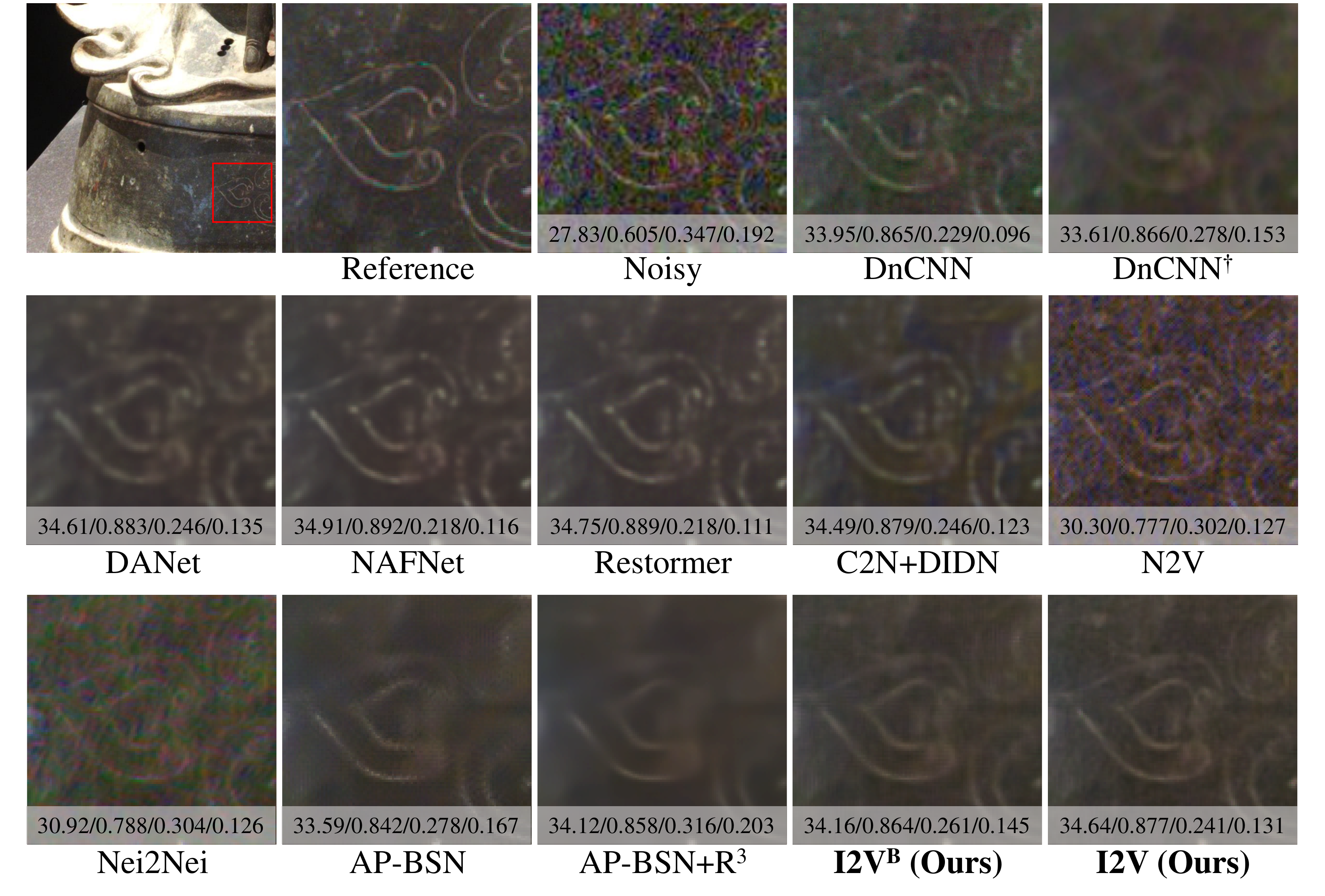}
\caption{Qualitative results for all comparison methods and our methods. Top is from NIND ISO3200. Bottom is from NIND ISO4000.}
\label{fig:figs5}
\end{figure*}

\begin{figure*}[t]
\centering
\includegraphics[width=15.5cm,keepaspectratio]{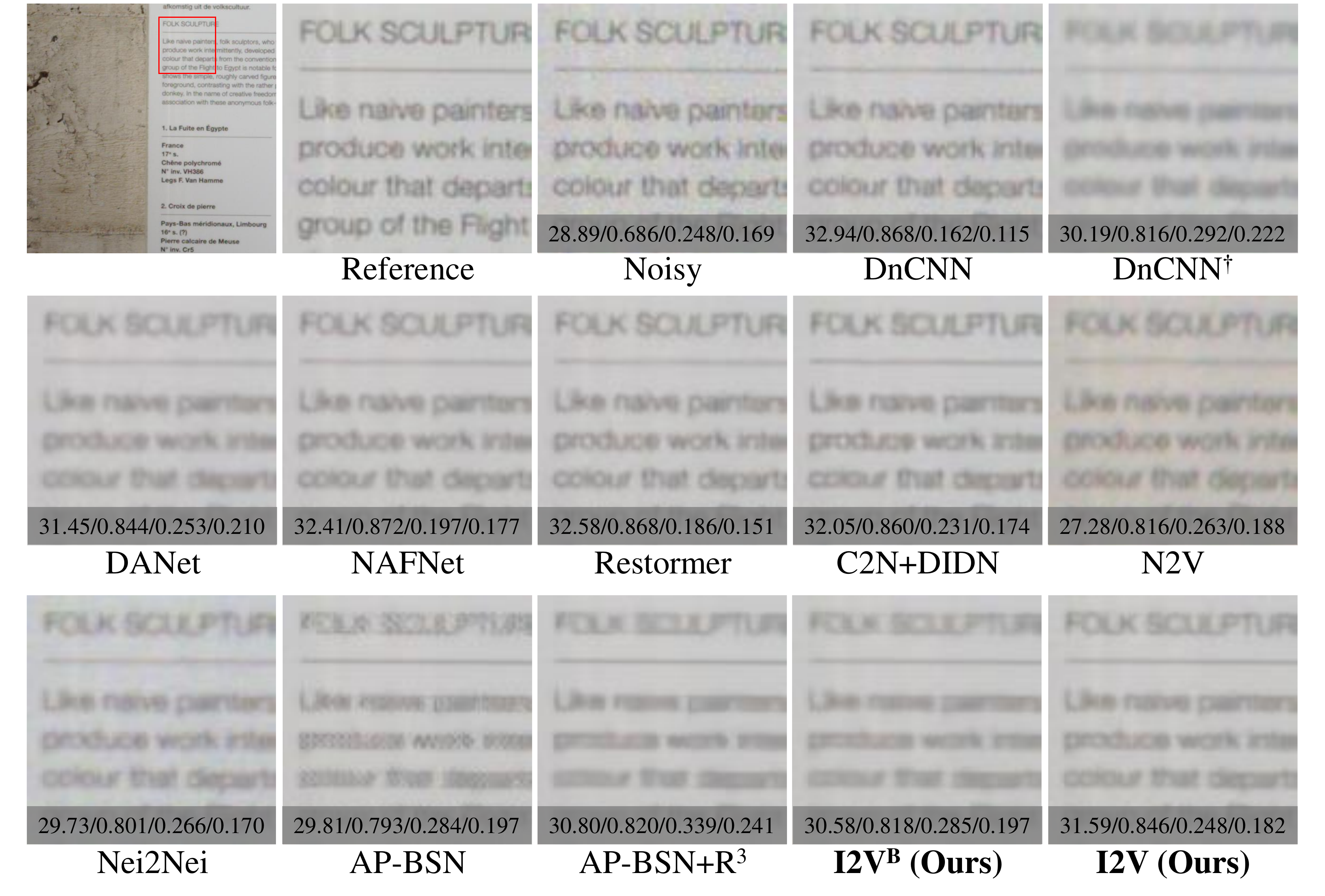}
\includegraphics[width=15.5cm,keepaspectratio]{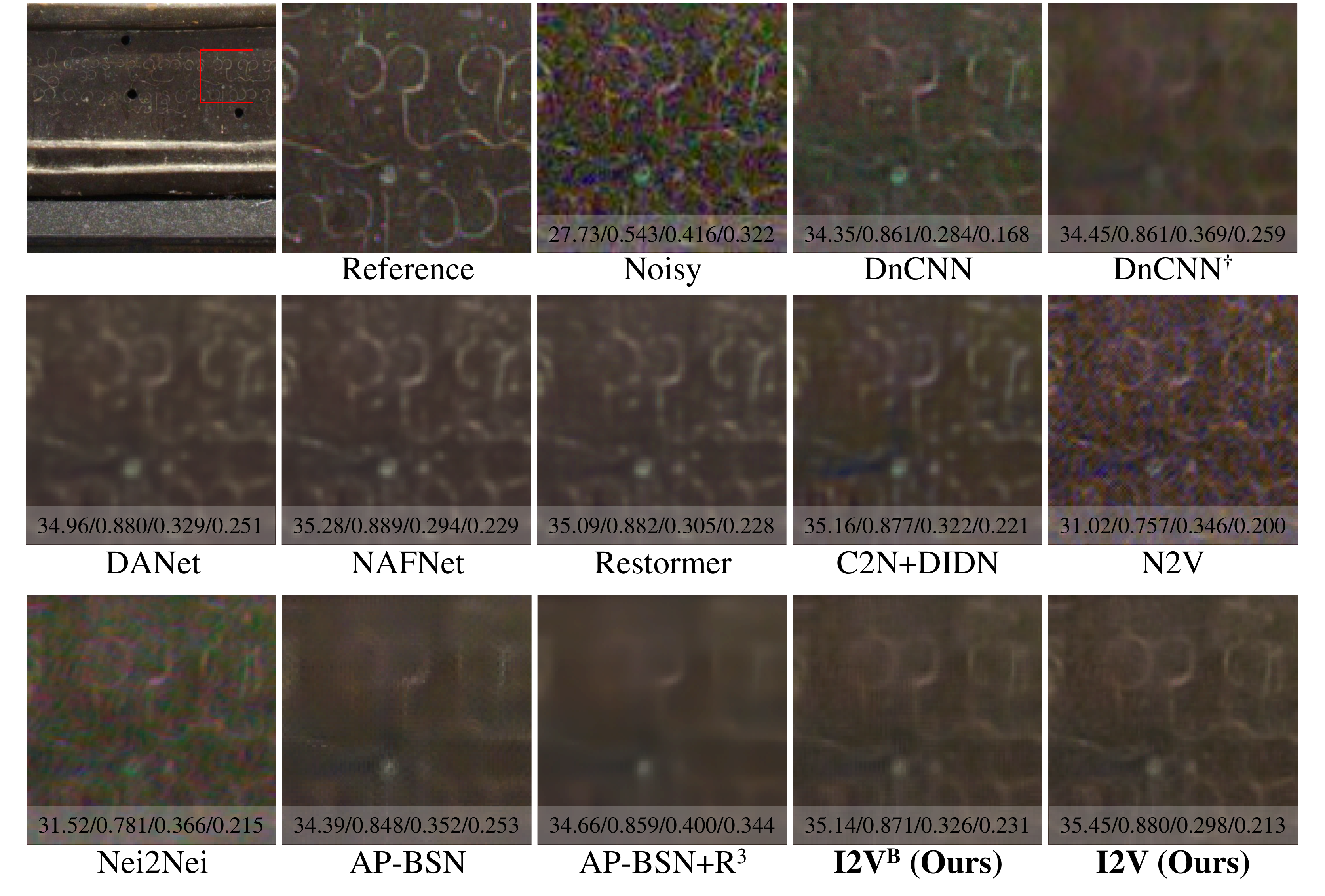}
\caption{Qualitative results for all comparison methods and our methods. Images are from the NIND ISO3200 dataset.}
\label{fig:figs6}
\end{figure*}

\begin{figure*}[t]
\centering
\includegraphics[width=15.5cm,keepaspectratio]{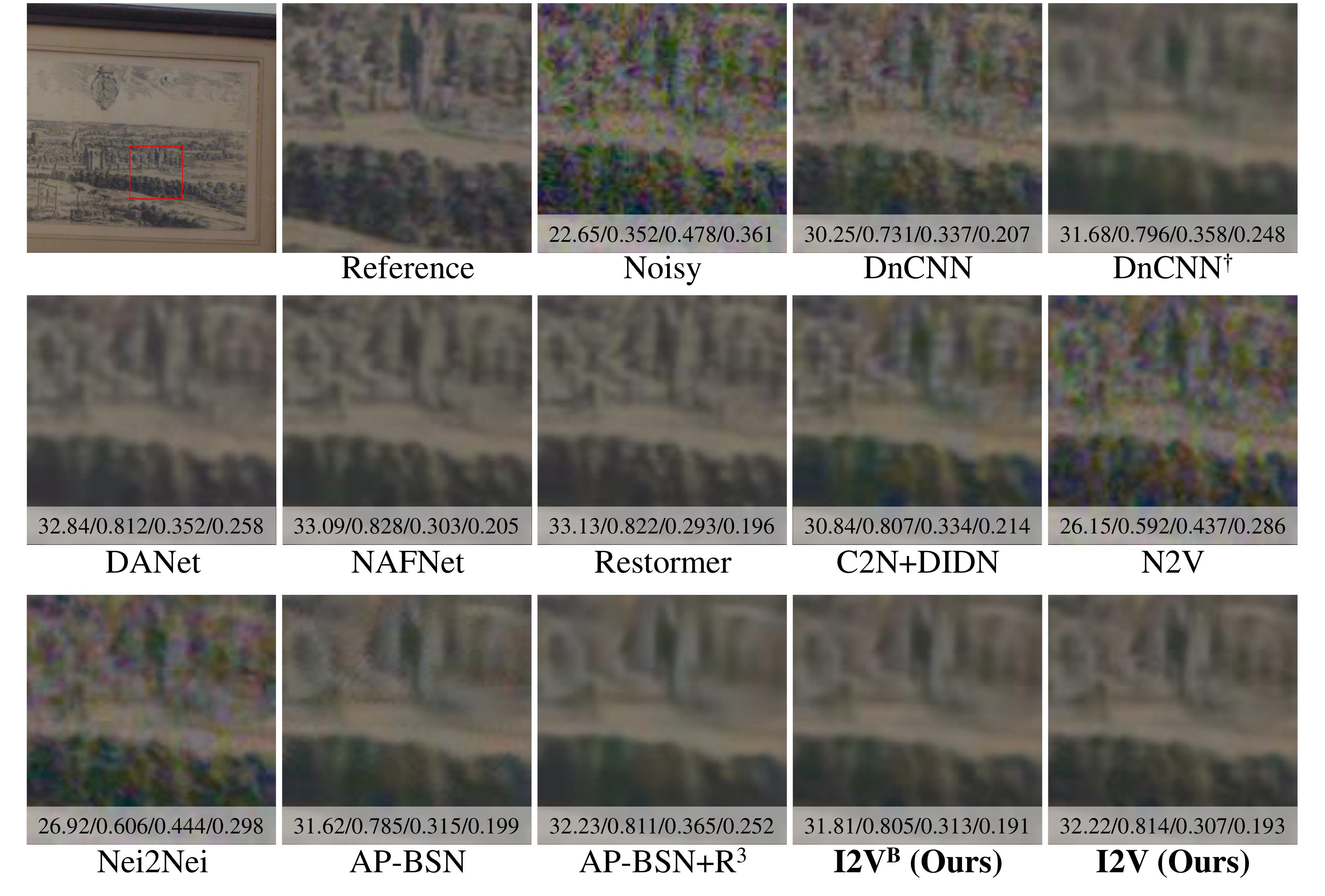}
\includegraphics[width=15.5cm,keepaspectratio]{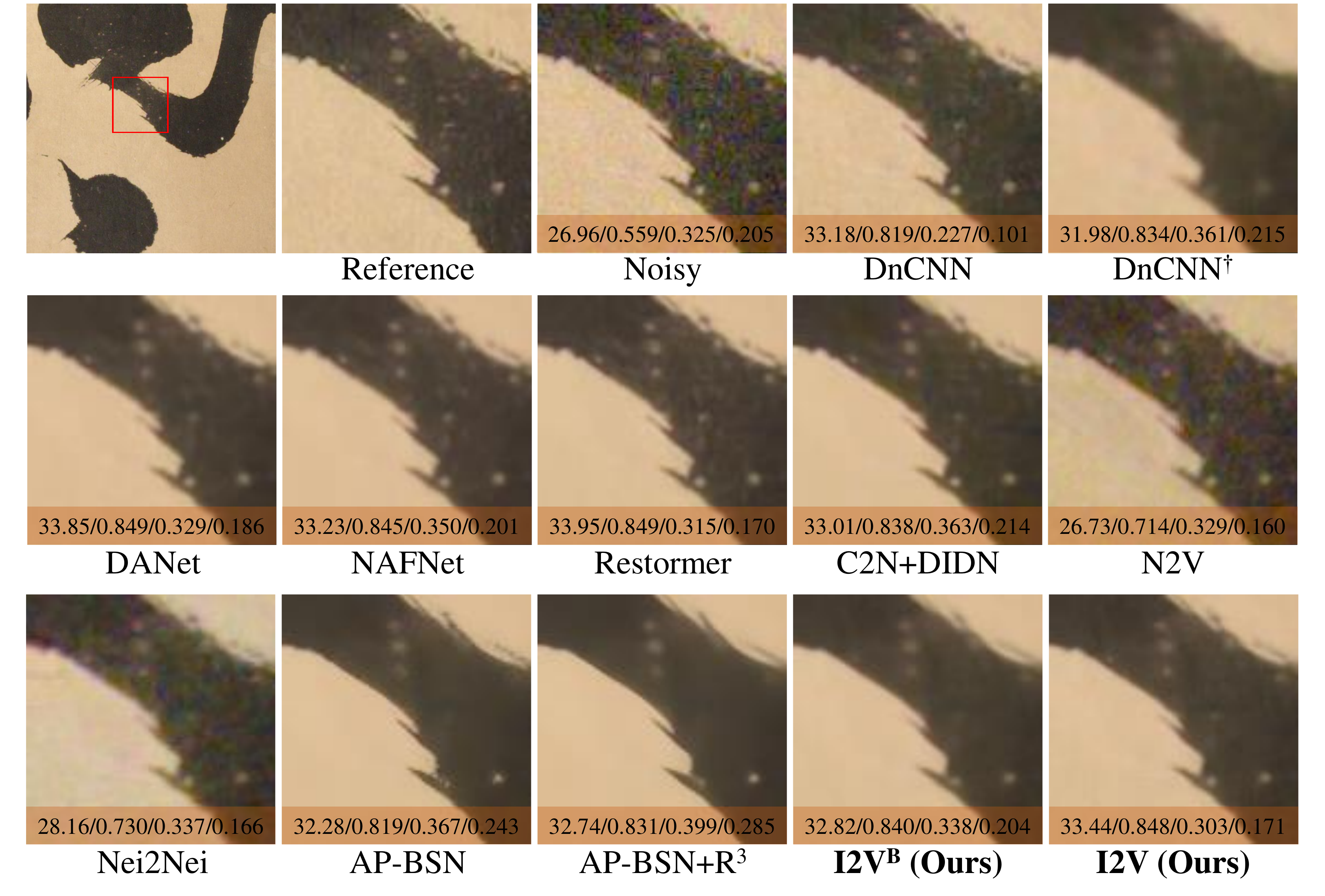}
\caption{Qualitative results for all comparison methods and our methods. Images are from the NIND ISO5000 dataset.}
\label{fig:figs7}
\end{figure*}

\begin{figure*}[t]
\centering
\includegraphics[width=15.5cm,keepaspectratio]{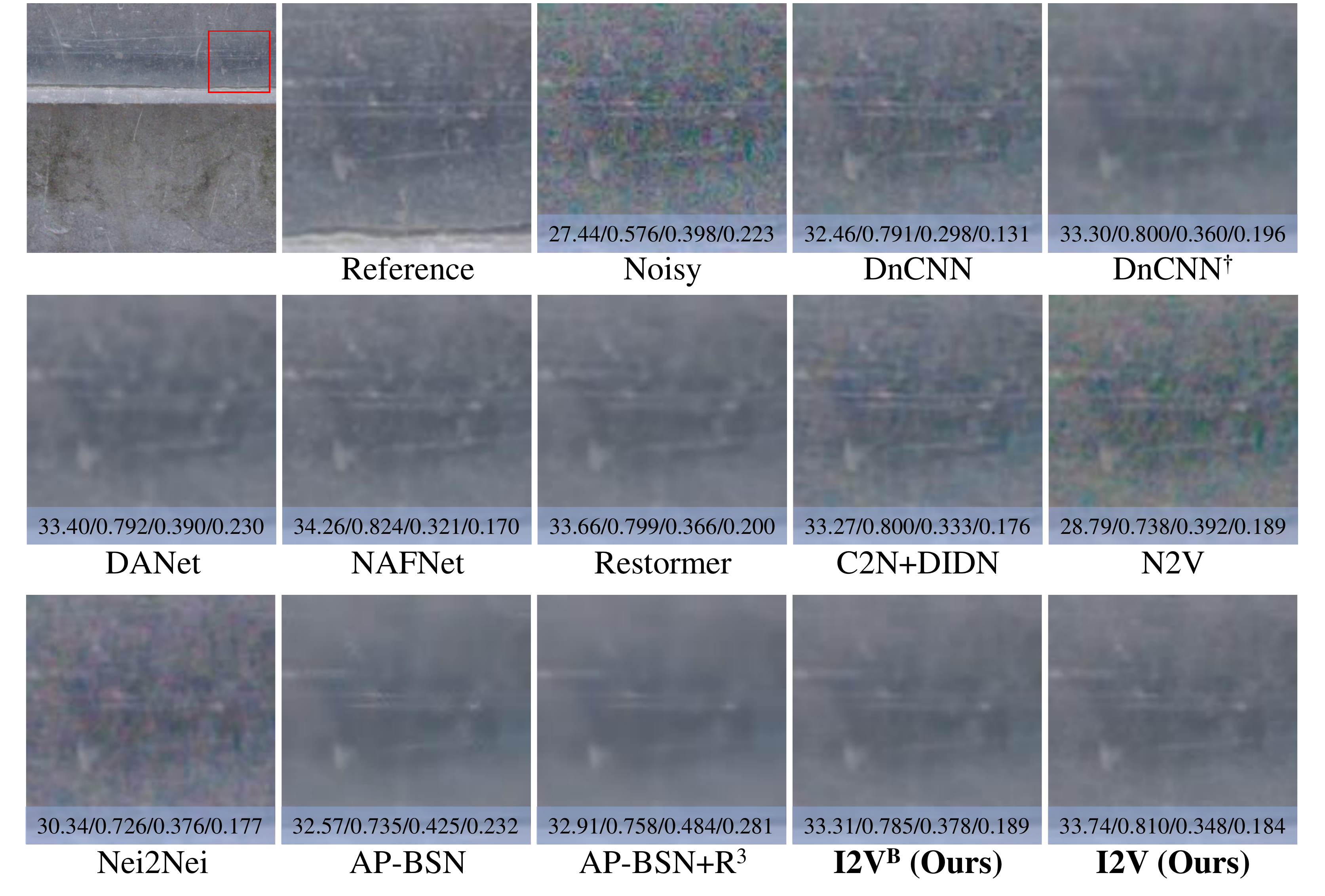}
\includegraphics[width=15.5cm,keepaspectratio]{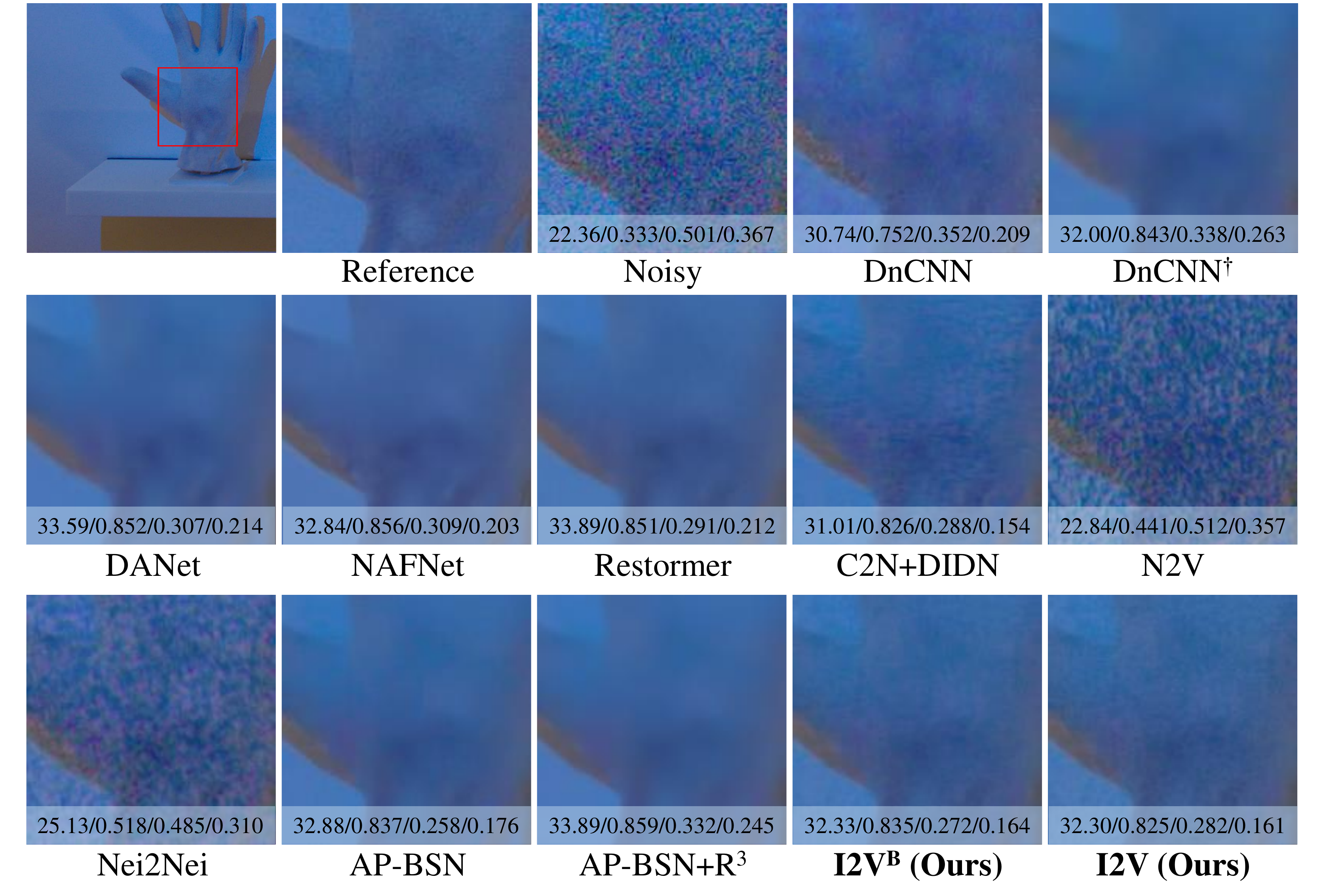}
\caption{Qualitative results for all comparison methods and our methods. Top is from the NIND ISO5000 dataset. Bottom is from the NIND ISO6400 dataset.}
\label{fig:figs8}
\end{figure*}

\begin{figure*}[t]
\centering
\includegraphics[width=15.5cm,keepaspectratio]{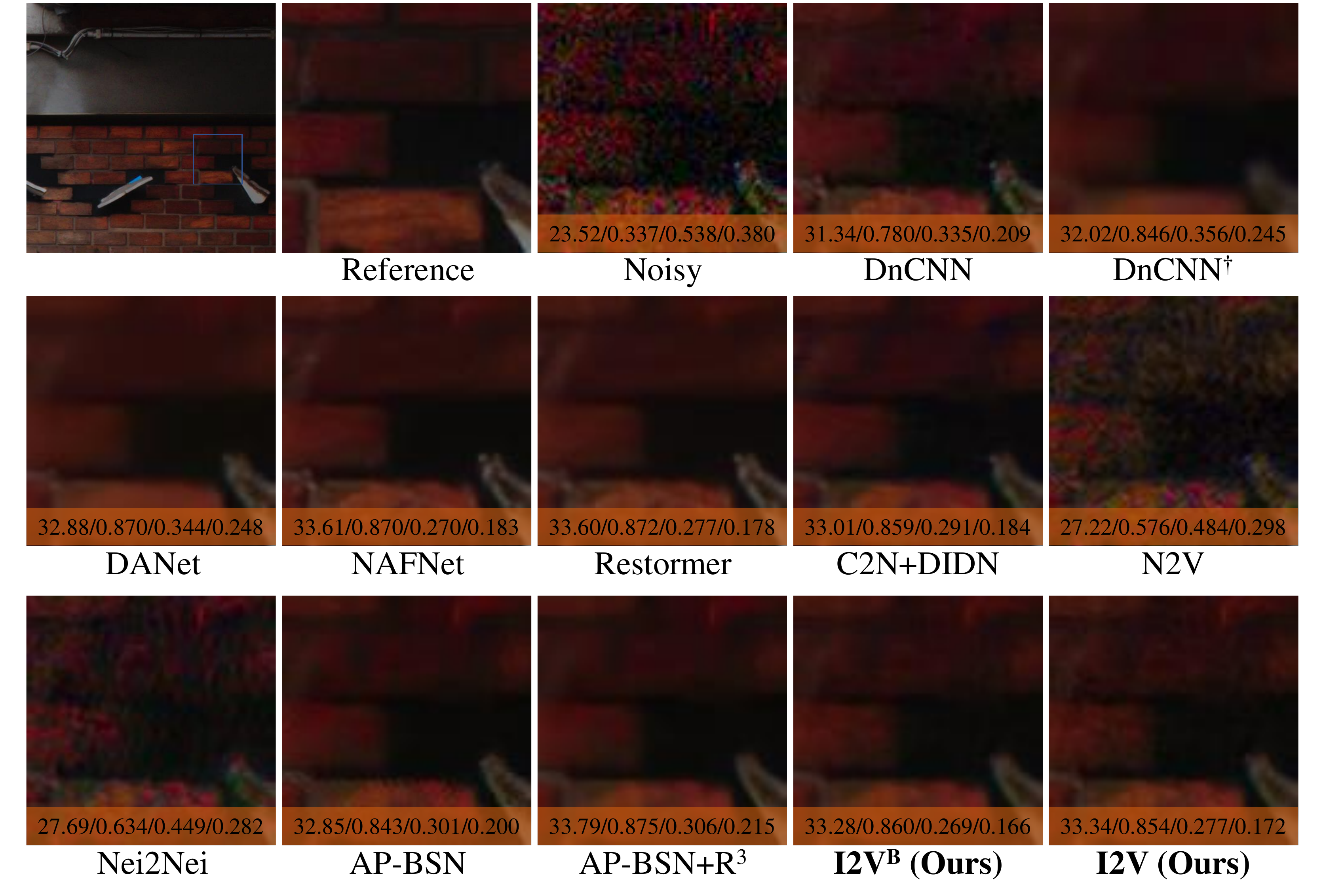}
\includegraphics[width=15.5cm,keepaspectratio]{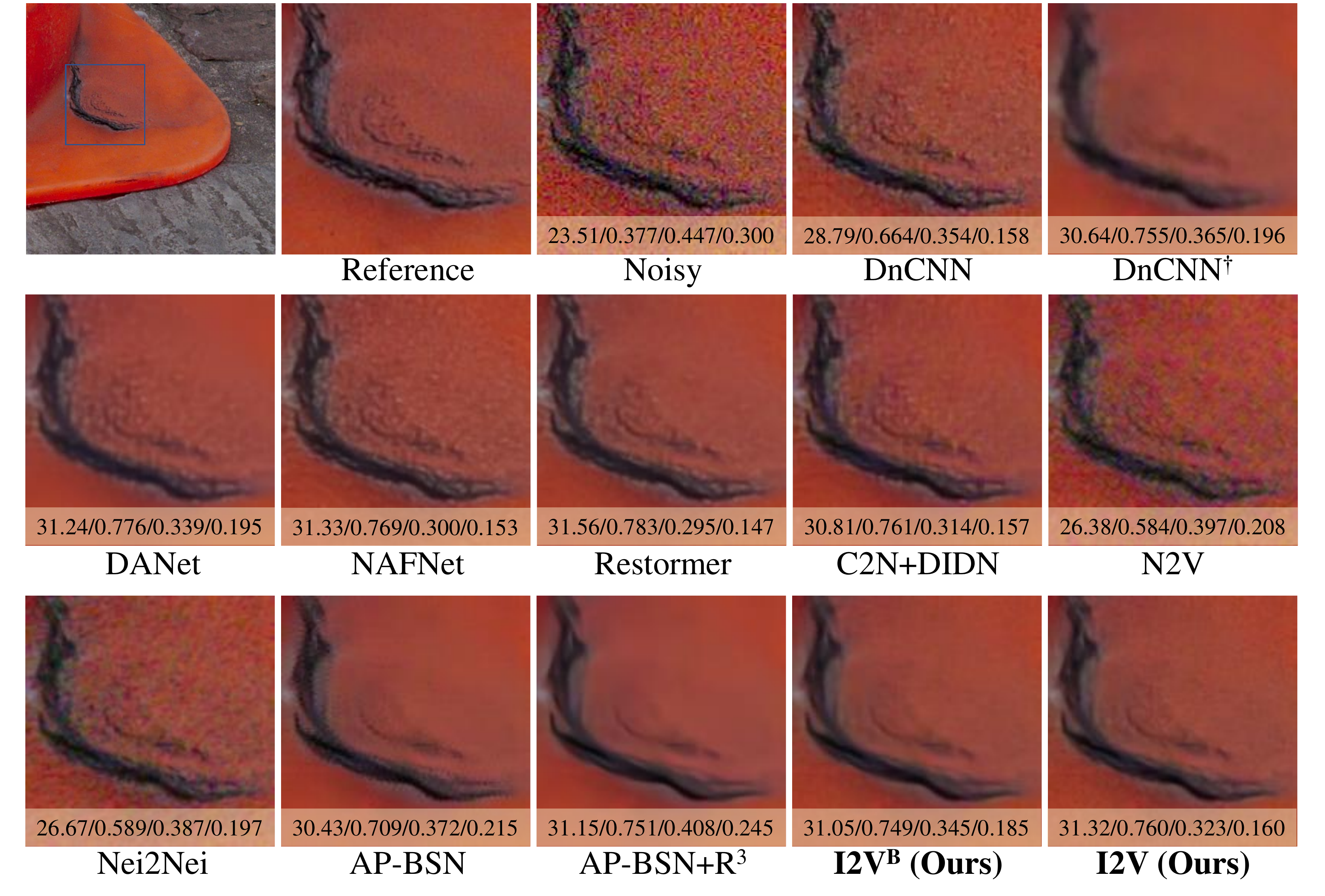}
\caption{Qualitative results for all comparison methods and our methods. Images are from the NIND ISO6400.}
\label{fig:figs9}
\end{figure*}

\begin{figure*}[t]
\centering
\includegraphics[width=15.5cm,keepaspectratio]{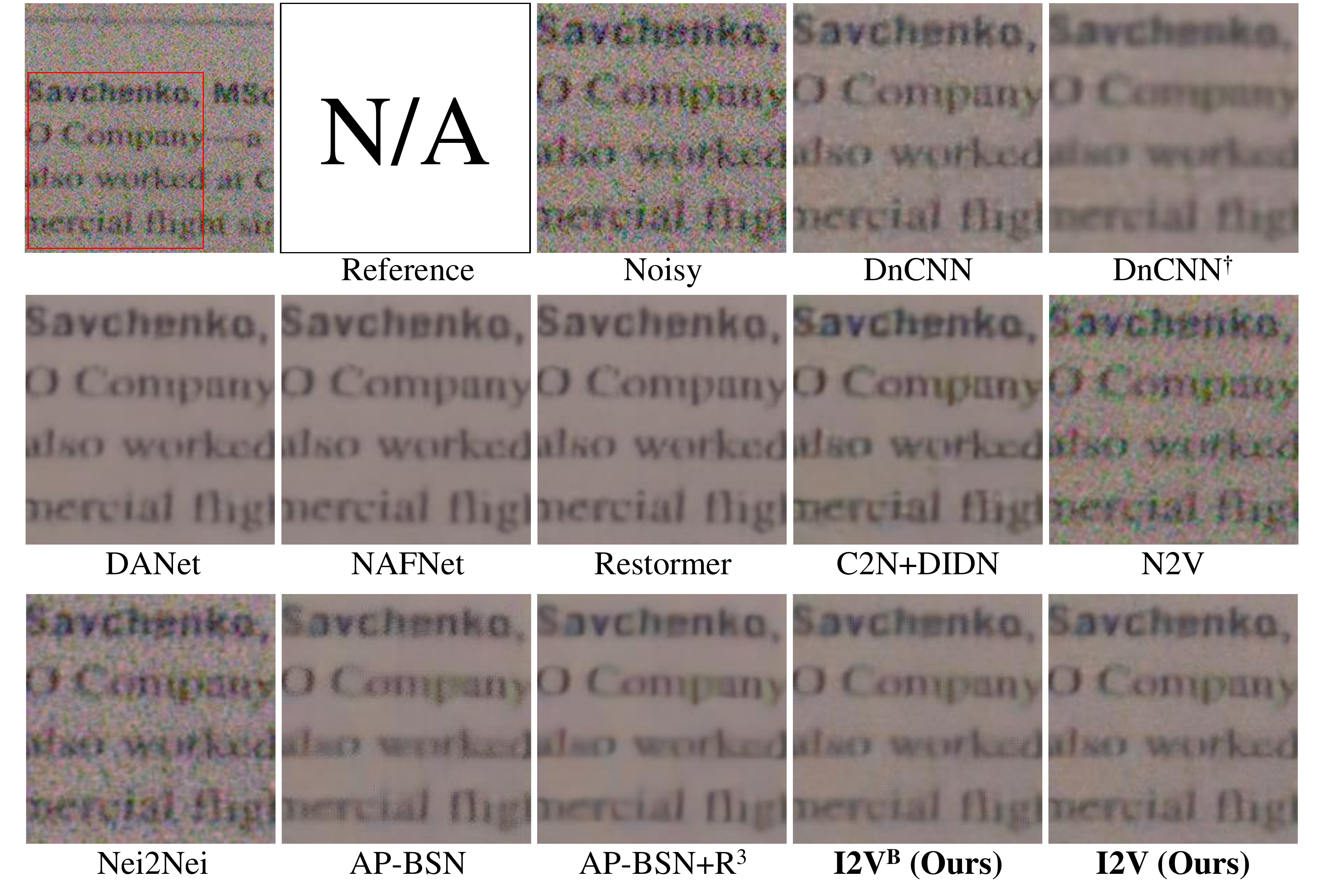}
\includegraphics[width=15.5cm,keepaspectratio]{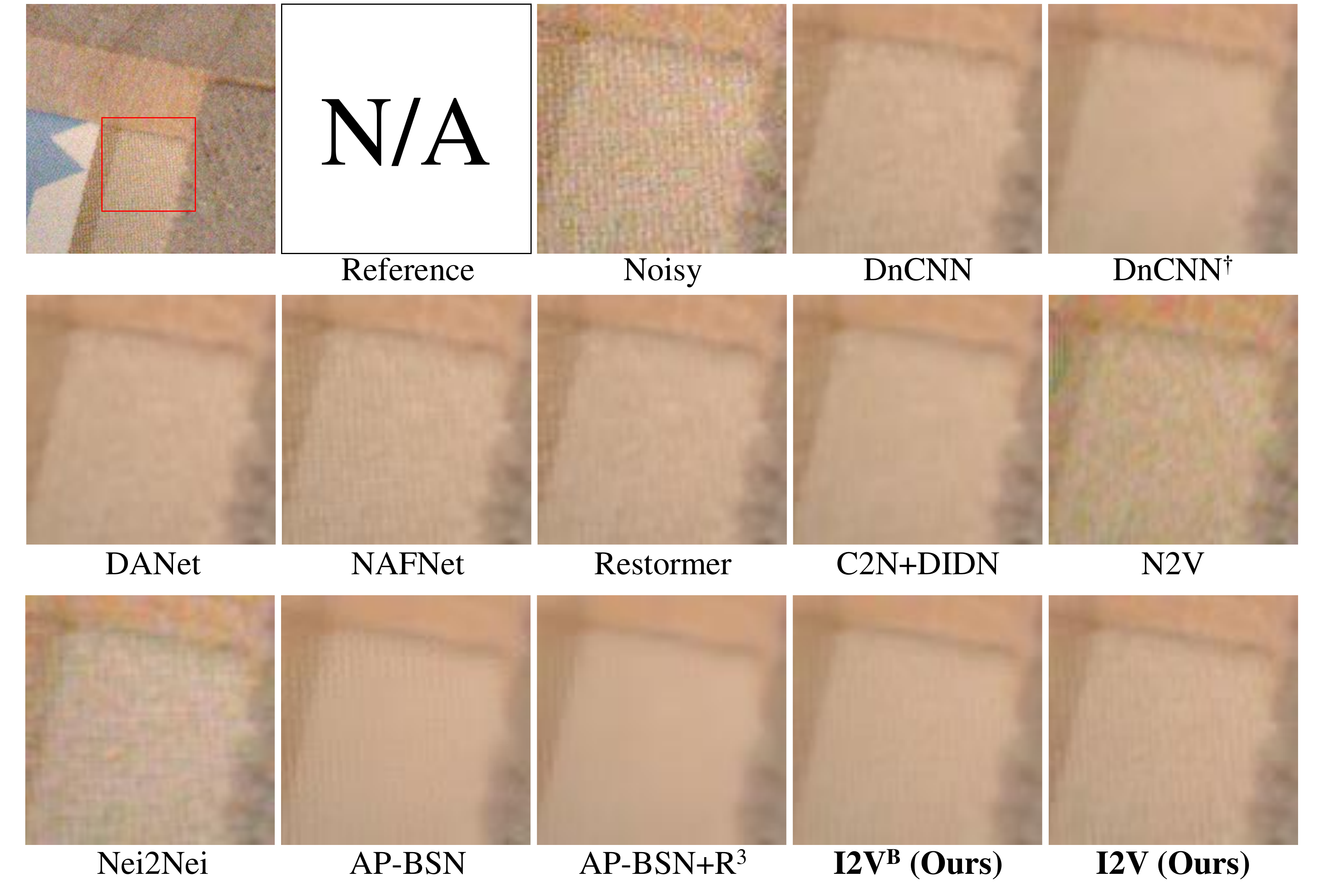}
\caption{Qualitative results for all comparison methods and our methods. Images are from the SIDD benchmark.}
\label{fig:figs10}
\end{figure*}

\begin{figure*}[t]
\centering
\includegraphics[width=15.5cm,keepaspectratio]{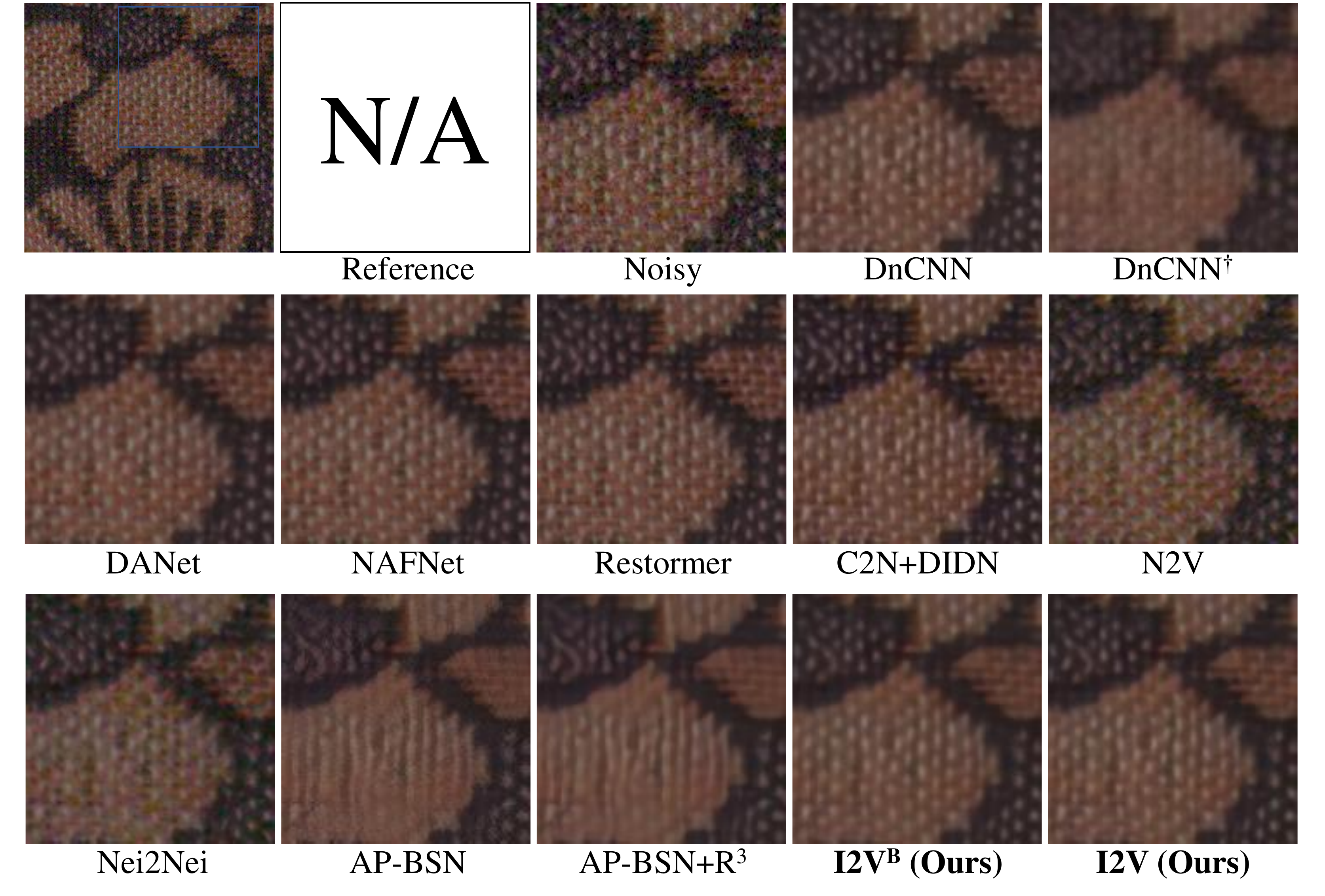}
\includegraphics[width=15.5cm,keepaspectratio]{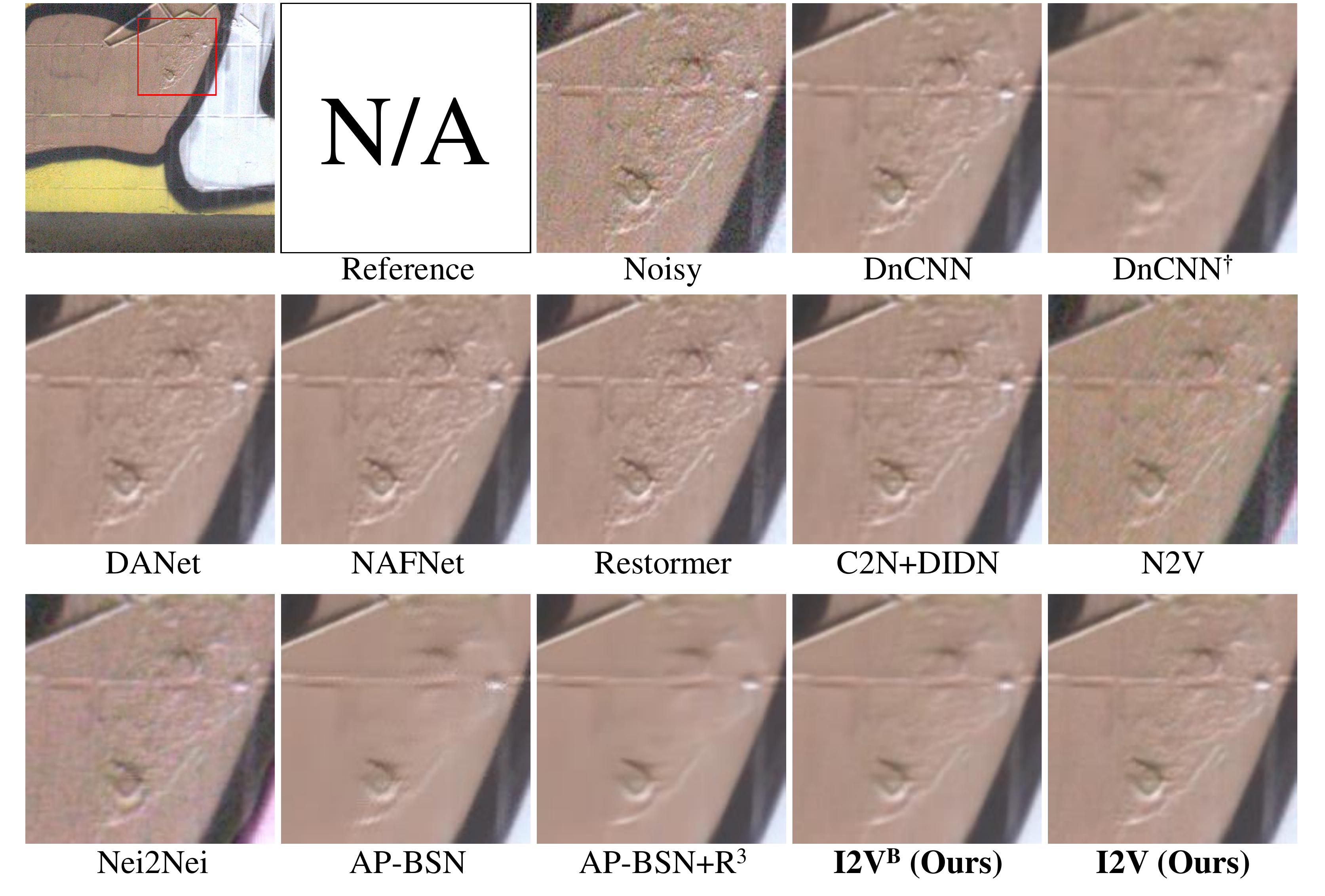}
\caption{Qualitative results for all comparison methods and our methods. Top is from the SIDD benchmark. Bottom is from the DND benchmark.}
\label{fig:figs11}
\end{figure*}

\begin{figure*}[t]
\centering
\includegraphics[width=15.5cm,keepaspectratio]{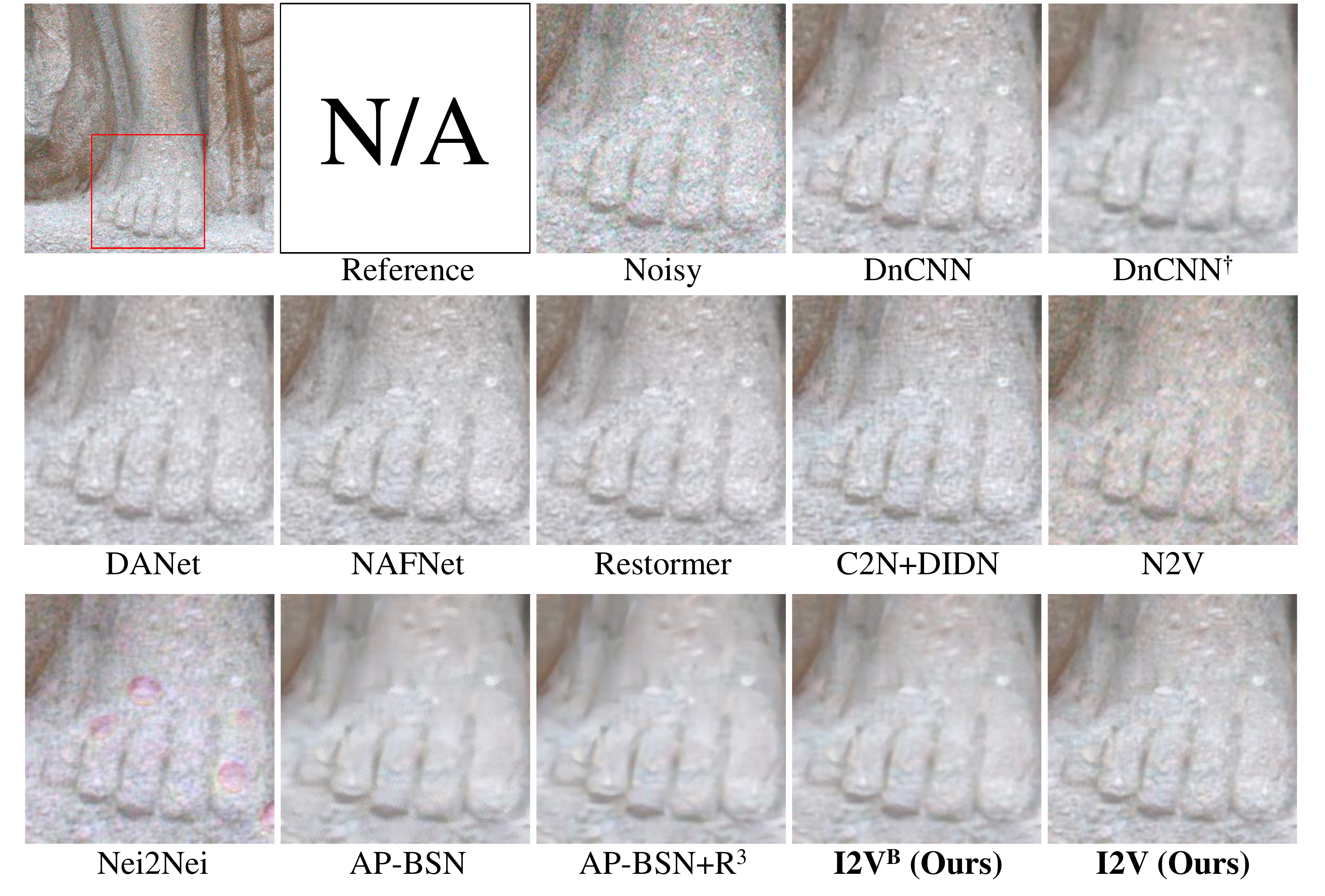}
\includegraphics[width=15.5cm,keepaspectratio]{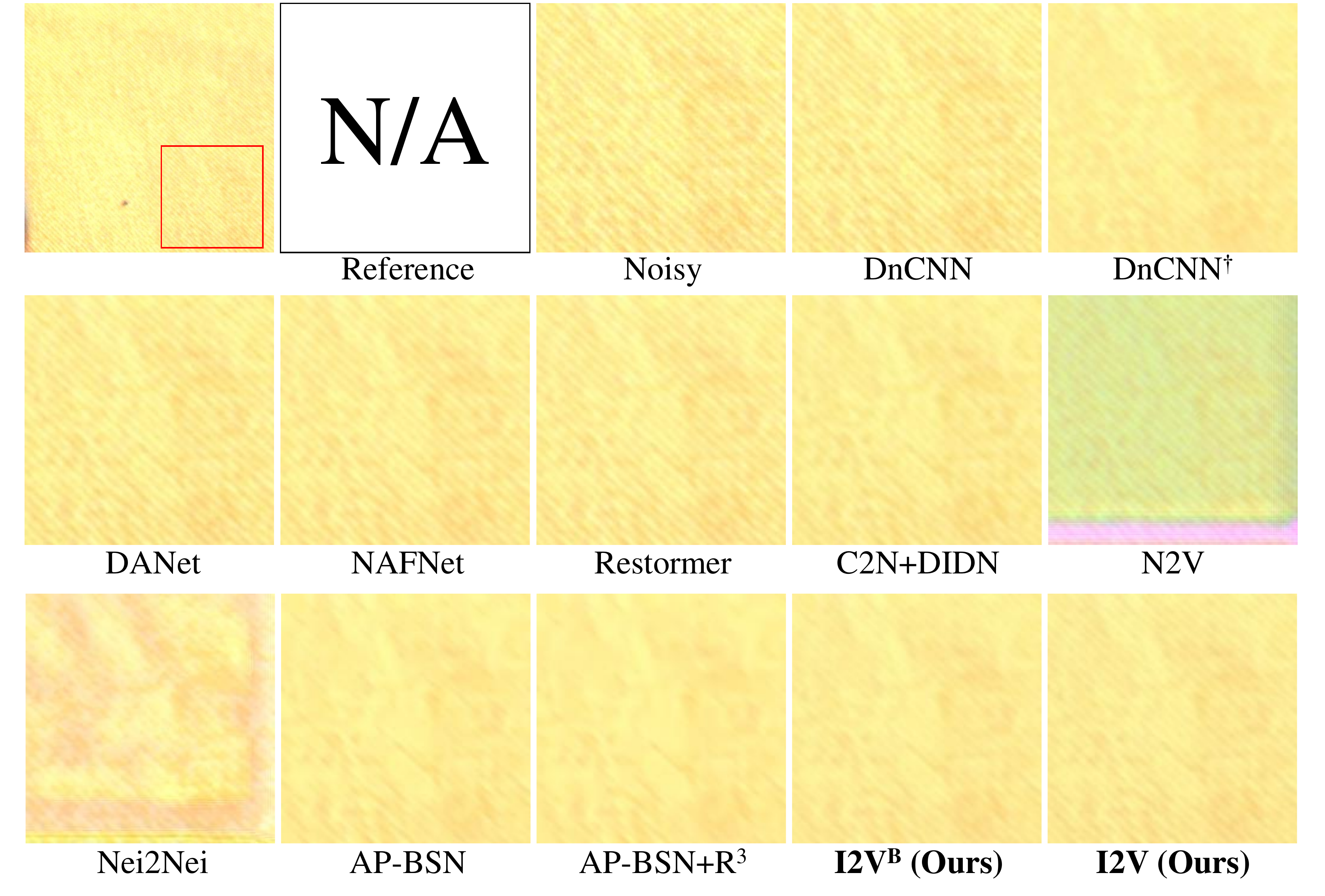}
\caption{Qualitative results for all comparison methods and our methods. Images are from the DND benchmark.}
\label{fig:figs12}
\end{figure*}

\section*{S4. More Quantitative and Qualitative Results}
In this section, we provide additional experiments results of the fully self-supervised denoising setting in the NIND dataset~\cite{nind} according to ISO3200 and ISO4000, as shown in Table~\ref{stab:tab3}. Moreover, we sample additional visual results for qualitative comparison for all methods used in our manuscript, as shown in Figure~\ref{fig:figs4} to \ref{fig:figs12}. Three images are selected on each dataset such as NIND ISO3200, NIND ISO4000, NIND ISO5000, NIND6400, SIDD~\cite{SIDD} benchmark, and DND~\cite{dnd} benchmark.
We denote N/A where its ground-truth is not available. From Figure~\ref{fig:figs4} to \ref{fig:figs12}, the performance under each sample from left to right indicates PSNR, SSIM, LPIPS, and DISTS, respectively. 
A left-topmost sample is an original clean image before zooming in. 
The notated performance is measured on the entire image (not on the zoomed-in region).
For the SIDD and DND benchmarks, we employed an original noisy image instead.



{\small
\bibliographystyle{ieee_fullname}
\bibliography{egbib}
}

\end{document}